\documentclass[twoside,11pt]{article} 

\pdfoutput=1 
\usepackage{jmlr2e_preprint}
\usepackage{bm}
\usepackage[yyyymmdd,hhmmss]{datetime}

\usepackage{url}
\usepackage{multirow}

\jmlrheading{1}{2000}{1-48}{4/00}{10/00}{Marina Meil\u{a} and Michael I. Jordan}

\ShortHeadings{Optimal Free Responses of GSFA and the Design of Training Graphs}{Escalante-B. and Wiskott}

\title{Theoretical Analysis of the Optimal Free Responses of Graph-Based SFA for the Design of Training Graphs (Preprint)}

\author{\name Alberto N. Escalante-B. \email alberto.escalante@ini.rub.de 
	\\ 
	\name Laurenz Wiskott \email laurenz.wiskott@ini.rub.de \\
        \addr Theory of Neural Systems \\
	Institut f\"ur Neuroinformatik \\ 
	Ruhr-University Bochum \\
	Bochum D-44801, Germany	
} 

\editor{Leslie Pack Kaelbling}

\usepackage[cmex10]{amsmath}
\usepackage{amssymb}
\usepackage{mathtools}

\providecommand{\myfloor}[1]{\left \lfloor #1 \right \rfloor }

\newcommand{\ignore}[1]{}
\newcommand*\Bell{\ensuremath{\boldsymbol\ell}}

\newcommand{\DDist}{\ensuremath{\mathcal{D}}}
\newcommand{\GG}{\ensuremath{\mathbf{\Gamma}}}
\newcommand{\MM}{\ensuremath{\mathbf{M}}}
\newcommand{\ELL}{\ensuremath{\ttxt{\tiny ELL}}}
\newcommand{\GGELL}{\ensuremath{\mathbf{\Gamma}^\ELL}}
\newcommand{\MMELL}{\ensuremath{\mathbf{M}^\ELL}}

\newcommand{\Lag}{\ensuremath{L}}
\newcommand{\e}{\ensuremath{\gamma}}
\newcommand{\Diag}{\ensuremath{\text{Diag}}}
\newcommand{\iif}{\ensuremath{\Leftrightarrow}}

\newcommand{\vect}[1]{\ensuremath{\mathbf{#1}}}

\newcommand{\Cov}{\ensuremath{\mathbf{C}}} 
\newcommand{\DCov}{\ensuremath{\mathbf{\dot{C}}}} 
\newcommand{\derivative}{derivative}

\newcommand{\todo}[1]{{\bf TODO:}#1\\
\line(1,0){200}

}
\renewcommand{\todo}[1]{}

\newcommand{\eqdef}{\ensuremath{\stackrel{\mathrm{def}}{=}}}

\newcommand{\eqlet}{\ensuremath{\stackrel{\mathrm{!}}{=}}}
\newcommand{\bigO}{\ensuremath{\mathcal{O}}}

\newcommand{\ttxt}[1]{\ensuremath{\textrm{#1}}}

\newcommand{\tcap}{} 

\graphicspath{{./svg_images/}}

\begin{document}

\maketitle
\begin{abstract}
Slow feature analysis (SFA) is an unsupervised learning algorithm that extracts slowly varying features from a time series.
Graph-based SFA (GSFA) is a \emph{supervised} extension that can solve regression problems if followed by a post-processing regression algorithm.
A training graph specifies arbitrary connections between the training samples.
The connections in current graphs, however, only depend on the \emph{rank} of the involved labels. Exploiting the exact label values makes further improvements in estimation accuracy possible.

In this article, we propose the exact label learning (ELL) method to create a graph that codes the desired label explicitly, so that GSFA is able to extract a normalized version of it directly.
The ELL method is used for three tasks: (1) We estimate gender from artificial images of human faces (regression) and show the advantage of coding additional labels, particularly skin color. (2) We analyze two existing graphs for regression. (3) We extract \emph{compact} discriminative features to classify traffic sign images. When the number of output features is limited, a higher classification rate is obtained compared to a graph equivalent to nonlinear Fisher discriminant analysis.
The method is versatile, directly supports multiple labels, and provides higher accuracy compared to current graphs for the problems considered.
\end{abstract} 


\section{Introduction}
The slowness principle is one of the learning paradigms that might explain, at least in part, the self-organization of neurons in the brain to extract invariant representations of relevant features.
This principle operates on an abstract level and postulates that the most relevant abstract information that can be extracted from the environment typically changes much slower than the individual sensory inputs (e.g., the position of a bug compared to the quickly changing neural activations in the retina of a frog observing it).

The slowness principle was probably first formulated by
\cite{Hinton-1989}, and online learning rules were developed shortly
after by \cite{Foldiak-1991} and \cite{Mitchison-1991}.
The first closed-form algorithm is referred to as slow feature analysis
\citep[SFA,][]{Wiskott-1998a,WisSej2002}.  


Recently, an extension of SFA for supervised learning called graph-based SFA~\citep[GSFA,][]{EscalanteWiskott-2013b} has been proposed.
In contrast to SFA, which is trained with a sequence of samples, GSFA is trained with a so-called training graph, in which the vertices are the samples and the edge weights represent similarities of the corresponding labels.
In SFA, slowness requires the minimization of the squared output differences between (temporally) consecutive pairs of samples, whereas in GSFA the pairs of samples do not have to be consecutive, are weighted, and are defined by the training graph.

GSFA can be used to explicitly exploit the available labels by establishing an output similarity objective involving arbitrary samples. Typically, GSFA is more effective than SFA at extracting a set of features that tend to concentrate the label information and allow the accurate prediction of the labels, implicitly solving the supervised learning problem.


Although GSFA and locality preserving projections~\citep[LPP,][]{HeNiyogi-2003} originate from different backgrounds and were first motivated with different goals and applications in mind, there is a close relation between them, sharing very similar objective functions and constraints. Two differences are that in GSFA the vertex weights are independent of the edge weights and that GSFA is invariant to the scale of the weights, providing a normalized objective function.
It is possible to use GSFA to compute LPP features, and vice versa. The results of this article might thus also be of interest for the LPP community.

\begin{figure}[bht!]
\begin{small}
\begin{center}
\includegraphics[width=0.95\textwidth]{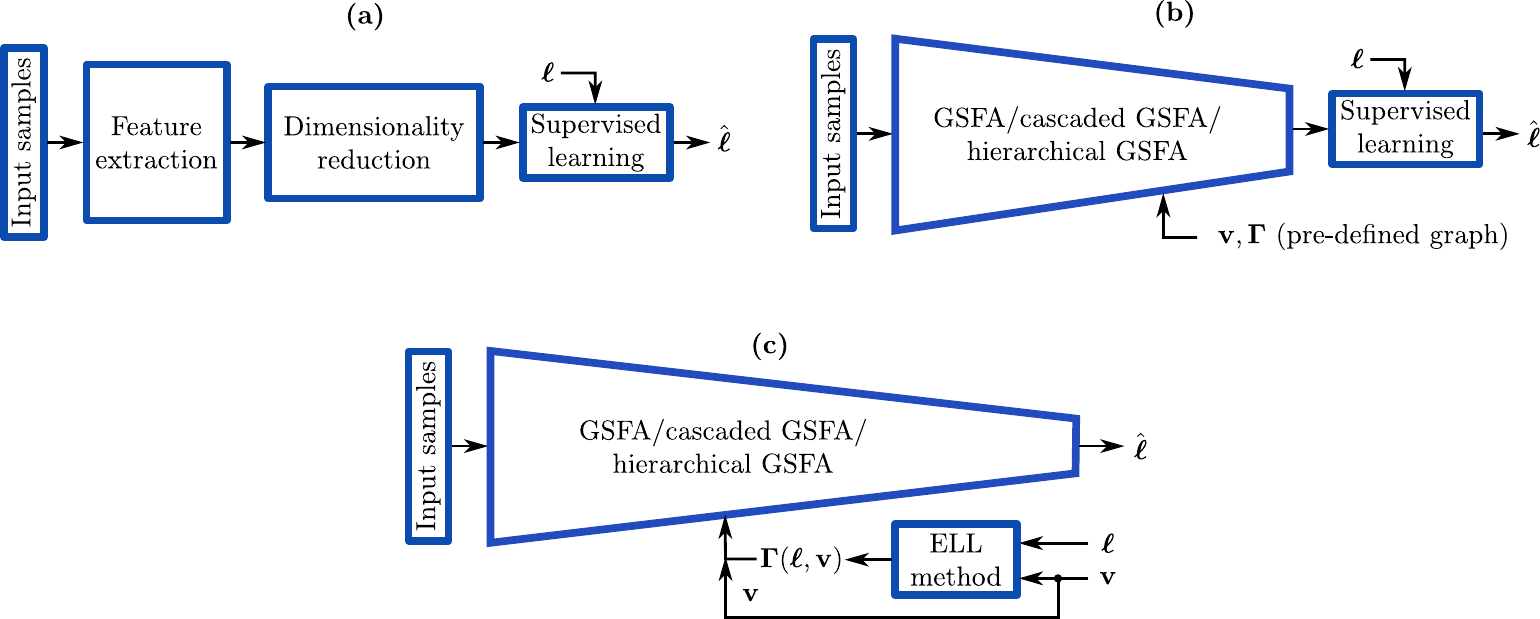}
\caption{
Three approaches for solving supervised learning problems. 
(a) A traditional and common approach. 
(b) Previous approach using GSFA with a pre-defined training graph, which is defined by the input samples, node weights $\vect{v}$, and edge weights $\GG$. The samples are assumed to be ordered by increasing label.
(c) The approach proposed here, which consists of a single GSFA architecture that is trained with a specially constructed graph $\GG(\Bell, \vect{v})$. 
The first slow feature (with a global sign adjustment) directly provides the label estimation.
If the label $\Bell$ does not have weighted zero mean and weighted unit variance, a final linear scaling should be included.}
\label{fig:ELLidea}
\end{center}
\end{small}
\end{figure}

In real life, many supervised learning problems are solved by applying feature extraction, (unsupervised) dimensionality reduction (DR), and an explicit supervised step (Figure~\ref{fig:ELLidea}.a).
Another approach (Figure~\ref{fig:ELLidea}.b), based on GSFA, first uses GSFA for \emph{supervised} DR, and then post-processes a small number of slow features with a conventional classification or regression algorithm. 
Supervised DR might result in higher accuracy than unsupervised DR.
The supervised learning problem is mostly solved by GSFA implicitly, because it frequently concentrates the label-predictive information in a few features.
Therefore, the post-processing step is not crucial, and might be a simple mapping from the slow features to the label domain.

Various training graphs for classification (clustered graph) and regression (e.g., serial, mixed, sliding window graphs) have been proposed~\citep{EscalanteWiskott-2013b}.
These pre-defined graphs offer great efficiency; although the number of edges contained in them is $\bigO(N^2)$, where $N$ is the number of samples, their structure makes the training complexity linear w.r.t. $N$.

However, the construction of pre-defined graphs only takes into account the rank of the labels and not their exact value, a simplification that might decrease the estimation accuracy.
In this article, we focus on the analysis and design of training graphs. We explore a new approach for solving regression problems with GSFA based on the construction of a special training graph, in which the slowest feature extracted is already a label estimation, up to a linear scaling (Figure~\ref{fig:ELLidea}.c).
To develop this exact label learning (ELL) method, we first study the slowest possible features that can be extracted by GSFA from a given graph when the feature space is unrestricted. 
Such features have also been called optimal free responses and have been computed for SFA in continuous time by \cite{Wiskott-2003a} using variational calculus. For GSFA, we use a different method based on linear algebra to cope with the discrete nature of the index $n$ that takes the place of the time.

Expressing the optimal free responses of GSFA in closed form then allows us to develop a theoretical method for the converse operation; from a set of free responses we design the corresponding training graph. 
The method allows the creation of a graph in which the slowest possible feature is the label to be learned.  
Moreover, one can learn \emph{multiple labels} simultaneously (e.g., object position, average color, shape, and size), and balance their importance.
This property can be exploited to learn \emph{auxiliary labels}, which provide a redundant coding of the original labels that may increase their estimation accuracy. 
In this case, as later explained, it is possible and desirable to emphasize the importance of the original labels over the auxiliary ones. 

Cascaded GSFA refers to the consecutive application of multiple passes of GSFA. One advantage over direct GSFA is that the joint feature space may be more complex.
Hierarchical GSFA (HGSFA), similarly to hierarchical SFA, is a divide-and-conquer approach for the extraction of slow features from high-dimensional data. HGSFA offers an excellent computational complexity compared to direct GSFA that can be as good as linear w.r.t the number of samples and the input dimensionality, depending on the network architecture.
A graph designed with the proposed ELL method can be used to train GSFA, cascaded GSFA, and HGSFA, being thus also applicable to high-dimensional data.

Although the ELL method is based on theory and mostly contributes to a deeper understanding of GSFA, it can also be used in practice, providing higher accuracy than pre-defined graphs. 
While in general the ELL method results in a higher complexity compared to GSFA trained with an efficient pre-defined graph, it is still computationally viable for some datasets without resorting to specialized hardware or parallelization. 

In the next section, we shortly review GSFA. In Section~\ref{sec:ELLforRegression}, we propose the ELL method. In Section~\ref{sec:applications_ELL}, we provide three applications.
Firstly, we solve a regression problem on gender estimation from artificial images, validating the method. 
Secondly, we analyze efficient pre-defined training graphs for regression.
Thirdly, we use the ELL method in a different way to design a training graph for the extraction of compact features for classification, yielding improved performance when $\log_2(C)$ to $C-2$ features are preserved.
For the first and third application, the accuracy is evaluated 
experimentally. Section~\ref{sec:discussion_ELL} closes the article with a discussion. 






\section{Graph-Based SFA (GSFA)}
\label{sec:gSFA} 
In this section, we recall the GSFA optimization problem, review the GSFA algorithm, 
and show how GSFA can be trained for both classification and regression.  

\subsection{Training Graphs and the GSFA Problem}
\label{introTrainingGraph}
GSFA is trained with a so-called training graph, in which the vertices are the samples and the edges between two samples may represent or be related to the similarity of their labels. 

\begin{figure}[htb!]
\begin{small}
\begin{center}
\includegraphics[width=0.9\textwidth]{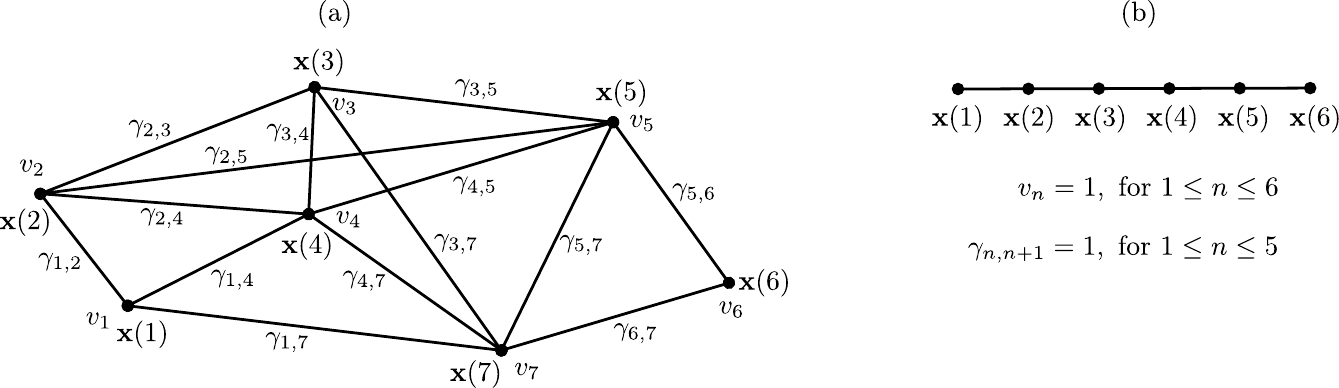}
\caption{{\tcap (a)} Example of a training graph with $N=7$ vertices. {\tcap (b)} A regular sample sequence (time series), which can be used to train SFA. This sequence is represented here as a linear graph that can be used with GSFA. If labels are available and the samples have been reordered by increasing/decreasing label (e.g., instead of having been ordered by time), it is called \emph{sample reordering} graph. \citep[Figure from][]{EscalanteWiskott-2013b}.}
\label{fig:training_graph}
\end{center}
\end{small}
\end{figure}

In mathematical terms, the training data is represented as a training graph
$G=(\vect{V},\vect{E})$ (illustrated in
Figure~\ref{fig:training_graph}.a) with a set $\vect{V}=\{\vect{x}(n)\}_n$ of vertices, each vertex being a sample,
and a set $\vect{E}$ of edges
$(\vect{x}(n), \vect{x}(n'))$, which are pairs of samples, with $1 \le
n,n' \le N$. 
The index $n$ (or $n'$) replaces the time variable $t$ used by SFA.
The edges are directed but typically have symmetric weights $\GG^T = \GG = \{\e_{n,n'} \}_{n',n}$;
weights $v_n > 0$ are associated with the vertices $\vect{x}(n)$ and can be used to reflect their importance, frequency, or reliability. 
This representation includes the standard time series of SFA as a special
case in which the graph has a linear structure (see Figure~\ref{fig:training_graph}.b).

Training graphs for classification typically favor connections between samples from the same class by means of larger edge weights compared to those of different classes, whereas training graphs for regression favor connections between samples with similar labels.

The concept of slowness has been generalized from sequences of samples (as in SFA) to training graphs.
The general goal is to extract features that fulfill certain normalization restrictions and minimize the sum of the weighted squared output differences of all connected samples.
More formally, the GSFA optimization problem~\citep{EscalanteWiskott-2013b}
can be stated as follows. For $1 \le j \le J$, find features $y_j(n) \eqdef g_j(\vect{x}(n))$, where $1 \le n \le N$ and $g_j$ belongs to the feature space $\mathcal{F}$ (frequent choices for $\mathcal{F}$ are all linear or quadratic transformations of the inputs), such that the objective function (weighted delta value)
\begin{equation} \label{eq:wobj_long}
\Delta_j \eqdef \frac{1}{R} \sum_{n,n'} \e_{n,n'} (y_j(n')-y_j(n))^2 \ttxt{ is minimal } 
\end{equation}
under the constraints
\begin{align} 
\label{eq:wzm_long}
\frac{1}{Q} \sum_{n} v_n y_j(n)  &= 0 \, ,\\
\label{eq:wuv_long}
\frac{1}{Q} \sum_{n} v_n (y_j(n))^2  &= 1 \, , \ttxt{ and} \\ 
\label{eq:wdec_long}
\frac{1}{Q} \sum_{n} v_n y_{j}(n) y_{j'}(n) &= 0 \, , \ttxt{ for }  j' < j \, ,  \\
\label{eq:R_Q_long}
\ttxt{with } Q \eqdef \sum_{n} v_n \;\;\;\;\; & \hspace{-1em}\ttxt{ and } \; R \eqdef \sum_{n,n'} \e_{n,n'} \, . 
\end{align}

The objective function penalizes the squared output differences between arbitrary pairs of samples using the edge weights as weighting factors.
The feature $y_1(n)$, for $1 \le n \le N$, is the slowest one, $y_2(n)$ is the second slowest, and so on. 
Constraints (\ref{eq:wzm_long})--(\ref{eq:wdec_long}) are called weighted zero mean, weighted unit variance, and weighted decorrelation, respectively.
They are similar to the normalization constraints of SFA, except for the inclusion of vertex weights. 
The factors $1/R$ and $1/Q$ are not essential for the optimization problem, but they provide invariance to the scale of the edge weights as well as to the scale of the vertex weights, and serve a normalization purpose.%

\subsection{Linear GSFA Algorithm} 
\label{sec:LGSFA}
The linear GSFA algorithm is similar to standard SFA~\citep{WisSej2002} and only differs in the computation of the matrices $\Cov$ and $\DCov$, which in GSFA takes into account the neighborhood structure specified by the training graph (samples, edges, and weights).
The sample covariance matrix $\Cov_{\vect{G}}$ is defined as: 
\begin{equation} \label{eq:cov_tg}
\Cov_{\vect{G}} \eqdef \frac{1}{Q} \sum_n v_n (\vect{x}(n) - \tilde{\vect{x}})(\vect{x}(n)- \tilde{\vect{x}})^T \,, 
\end{equation}
where $\vect{x}(n)$ and $v_n$ denote an input sample and its weight, respectively, and 
\begin{equation}
  \tilde{\vect{x}} \,\eqdef\, \frac{1}{Q} \sum_n v_n \vect{x}(n) \label{eq:hatX}
\end{equation}
is the weighted average of all samples.
The \derivative\ second-moment matrix $\DCov_{\vect{G}}$ is defined as:
\begin{equation} \label{eq:dcov_tg}
\DCov_{\vect{G}} \eqdef \frac{1}{R} \sum_{n,n'} \e_{n,n'} \big(\vect{x}(n') - \vect{x}(n)\big) \big(\vect{x}(n') - \vect{x}(n)\big)^T \,,
\end{equation}
where edge weights $\e_{n,n'}$ are defined as $0$ if the graph does not have an edge $(\vect{x}(n), \vect{x}(n'))$.  

Given these matrices, a sphering matrix $\vect{S}$ and a rotation matrix $\vect{R}$ are computed with 
\begin{align} \label{eq:sphering_C_g}
\vect{S}^T \Cov_{\vect{G}} \vect{S} \, &= \, \vect{I} \, , \, \ttxt{ and} \\
\label{eq:sphering_DC_g}
\vect{R}^T \vect{S}^T \DCov_{\vect{G}} \vect{S} \vect{R} \, &= \, \bm{\Lambda} \, ,
\end{align}
where $\bm{\Lambda}$ is a diagonal matrix with diagonal elements $\lambda_1 \le \lambda_2 \le \cdots \le \lambda_J$.   
Finally the algorithm returns $\Delta(y_1), \dots, \Delta(y_J)$, $\vect{W}$ and $\vect{y}(n)$, where
\begin{align} \label{eq:W_is_SR}
\vect{W} \, &= \, \vect{S} \vect{R} \, , \, \ttxt{ and} \\
\label{eq:y_tg}
\vect{y}(n) \, &= \, \vect{W}^T (\vect{x}(n) - \tilde{\vect{x}}) \, . 
\end{align}

It has been shown that the GSFA algorithm presented above indeed solves the optimization problem (\ref{eq:wobj_long}--\ref{eq:wdec_long}) in the linear function space. The proof is similar to the corresponding proof of standard linear SFA~\citep{WisSej2002}.

\paragraph{Probabilistic interpretation of a graph.}
Interestingly, if the graph is connected and the following consistency restriction is fulfilled
\begin{align}
\label{eq:consistency_long}
\forall n: v_n/Q = \sum_{n'}\e_{n,n'}/R ,
\end{align}
then GSFA yields the same features as standard SFA trained on a sequence generated by using the graph as a Markov chain with transition probabilities $\e_{n,n'}/R$~\citep[see][]{KlampflMaass:09b,EscalanteWiskott-2013b}.
Thus, one can use SFA to emulate GSFA. However, depending on the training graph used, emulating GSFA with SFA may be more computationally expensive. 

\subsection{Clustered and Serial Training Graphs}
\label{sec:clustered_serial}
In this section, we present two efficient pre-defined graphs, the \emph{clustered graph} for classification and the \emph{serial graph} for regression~\citep{EscalanteWiskott-2013b}.

\begin{figure}[ht!]
\begin{small}
\begin{center}
\includegraphics[width=0.5\columnwidth]{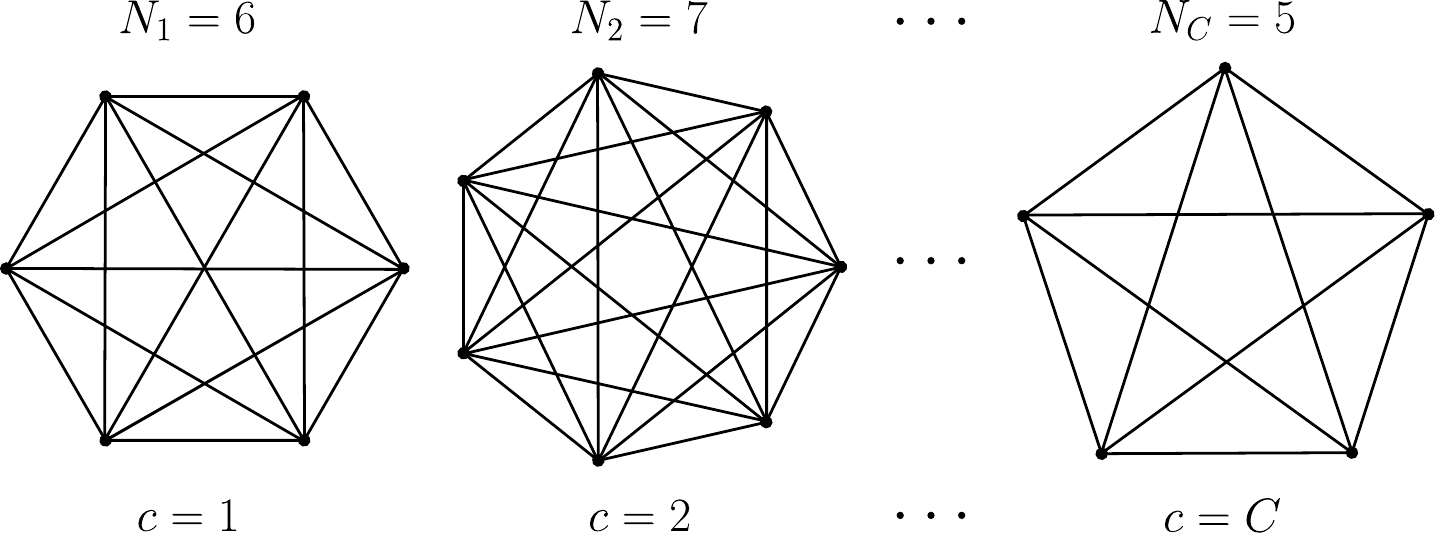} 
\caption{Illustration of a \emph{clustered} training graph used for a classification problem with $C$ classes. Each vertex represents a sample, and edges represent transitions.
The $N_c$ samples belonging to a class $c \in \{1, \dots, C\}$ are connected, constituting a fully connected subgraph. Samples of different classes are not connected. 
Vertex weights are identical and equal to one, whereas edge weights depend on the cluster size as $\e_{n,n'} = 1/(N_c-1)$, where $\vect{x}(n)$ and $\vect{x}(n')$ belong to class $c$ and $n \neq n'$. \citep[Figure from][]{EscalanteWiskott-2013b}.}
\label{fig:complete_training}
\end{center}
\end{small}
\end{figure}

The clustered graph, described in Figure~\ref{fig:complete_training}, generates features useful for classification.
The optimization problem associated with this graph explicitly demands that samples from the same class should typically be mapped to similar outputs.

The features learned by GSFA on this graph are equivalent to those learned by 
Fisher discriminant analysis (FDA, see \citealp{KlampflMaass:09b} and also compare~\citealp{Berkes-2005a} and~\citealp{Berkes05}). 
This type of problem can be analyzed theoretically when the function space of SFA is unrestricted. Consistent with FDA, the first $C-1$ slow features extracted (optimal free responses) are orthogonal step functions, and are piece-wise constant for samples from the same class~\citep{Berkes-2005a}. 

\begin{figure}[ht!]
\begin{small}
\begin{center}
\includegraphics[width=0.40\columnwidth]{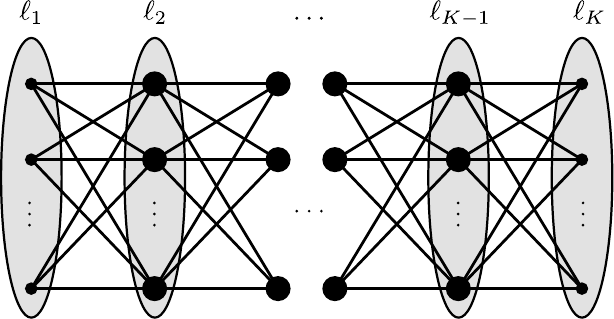} 
\caption{Illustration of a serial training graph with $K$ discrete label values $\{ \ell_1, \ldots, \ell_{K} \}$. In this graph, the vertices are ordered and then grouped according to their discrete label. Even though the original label of two samples might differ, they will be grouped together if they have the same discrete label. Each dot represents a sample, edges represent connections, and ovals represent groups of samples. 
The samples of groups with discrete labels $\ell_2$ to $\ell_{K-1}$ have a weight of $2$, whereas the samples of extreme groups with labels $\ell_1$ or $\ell_{K}$ have a weight of $1$ (sample weights represented by bigger/smaller dots). The weight of all edges is $1$. 
\citep[Figure from][]{EscalanteWiskott-2013b}.}
\label{fig:serial_training_graph}
\end{center}
\end{small}
\end{figure}

The \emph{serial} training graph, described in Figure~\ref{fig:serial_training_graph}, is constructed by discretizing the original label $\Bell$ into a relatively small set of $K$ discrete label values, namely $\{ \ell_1, \ldots, \ell_{K} \}$, where $\ell_1 < \ell_2 < \cdots < \ell_{K}$.
Afterwards, the samples are divided into $K$ groups of size $N/K$ sharing the same discrete labels.
Edges connect all pairs of samples from consecutive groups with discrete labels $\ell_{k}$ and $\ell_{k+1}$, for $1 \le k \le K-1$. Thus, connections are only inter-group, and intra-group connections do not appear.
Notice that since any two vertices of the same group are adjacent to exactly the same neighbors, they are likely to be mapped to similar outputs by GSFA.
Following GSFA a complementary explicit regression step on a few features solves the original regression problem.
There are several efficient graphs for regression besides the \emph{serial graph}. We employ the serial graph in this article, because it has consistently given good results for various regression problems. 

The clustered and serial graphs allow efficient training in linear time w.r.t $N$, whereas the number of connections considered is $\bigO(N^2)$ if the number of clusters or groups is constant.

\subsection{GSFA Optimization Problem in Matrix Notation}
In order to apply linear algebra methods to analyze GSFA, we use matrix notation. In what follows we assume that the edge weights are symmetric\footnote{An asymmetric edge-weight matrix $\GG$ can be converted into a symmetric one $\GG' \eqdef \frac{\GG+\GG^T}{2}$ without altering the solution to the optimization problem.} $(\GG = \GG^T)$ and that the consistency restriction (\ref{eq:consistency_long}) is fulfilled. This restriction can also be written as 
\begin{align}
\label{eq:consistency}
\vect{v} \stackrel{(\ref{eq:consistency_long})}{=} \frac{Q}{R} \GG \vect{1} \, ,
\end{align}
where $\vect{1}$ is a vector of ones of length $N$. 

If $\vect{y}$ is a feasible solution (i.e., satisfying (\ref{eq:wzm_long}) and (\ref{eq:wuv_long})) and the graph fulfills the consistency restriction (\ref{eq:consistency}), the weighted delta value (\ref{eq:wobj_long}) can be simplified as follows,
\begin{eqnarray}
\Delta_{\vect{y}} & \stackrel{(\ref{eq:wobj_long})}{=} & \frac{1}{R} \sum_{n,n'} \e_{n,n'} (y(n')-y(n))^2  \\
 &=& \frac{1}{R} \Big( \sum_{n'} (y(n'))^2 \sum_{n} \e_{n,n'} + \sum_{n} (y(n))^2 \sum_{n'} \e_{n,n'}  - 2 \sum_{n,n'} \e_{n,n'} y(n') y(n) \Big) \\
 &\stackrel{(\ref{eq:consistency})}{=}& \frac{1}{R} \Big( \sum_{n'} (y(n'))^2 \frac{R}{Q} v(n') + \sum_{n} (y(n))^2 \frac{R}{Q} v(n)  - 2 \vect{y}^T \GG \vect{y} \Big) \\
\label{eq:wobj_simp}
&\stackrel{(\ref{eq:wuv_long})}{=}& 2 - \frac{2}{R} \vect{y}^T \GG \vect{y} \, .
\end{eqnarray}

The optimization problem can then be stated as: For $1 \le j \le J$, find vectors $\vect{y}_j$ of length $N$, with $y_j(n) \eqdef g_j(\vect{x}(n))$ and $g_j \in \mathcal{F}$, minimizing 
\begin{align}
\label{eq:wobj}
\Delta_j \, &\stackrel{(\ref{eq:wobj_long},\ref{eq:wuv_long},\ref{eq:consistency})}{=} \, 2 - \frac{2}{R} \vect{y}_j^T \GG \vect{y}_j
\end{align}
subject to:
\begin{align}
\label{eq:wzm}
\vect{v}^T \vect{y}_j &\stackrel{(\ref{eq:wzm_long})}{=} 0 \\
\label{eq:wuv}
\vect{y}_j^T \Diag(\vect{v}) \vect{y}_j &\stackrel{(\ref{eq:wuv_long})}{=} Q \\
\label{eq:wdec}
\vect{y}_j^T \Diag(\vect{v}) \vect{y}_{j'} &\stackrel{(\ref{eq:wdec_long})}{=} 0 , \ttxt{ for }  j' < j \, ,
\end{align}
where 
\begin{align}
\label{eq:Q}
Q &\stackrel{(\ref{eq:R_Q_long}.a)}{=} \vect{1}^T \vect{v} \, , \\
\label{eq:R}
R &\stackrel{(\ref{eq:R_Q_long}.b)}{=} \vect{1}^T \GG \vect{1} \, ,
\end{align}
and $\Diag(\vect{v})$ denotes a diagonal matrix with diagonal $\vect{v}$.

\section{Explicit Label Learning for Regression Problems}
\label{sec:ELLforRegression}
\subsection{Optimal Free Responses of GSFA}
\label{sec:critical_points}
In this section, we calculate the slowest possible solutions (optimal free responses) to the GSFA problem (\ref{eq:wobj})--(\ref{eq:wdec}) that one could find if the feature space were unlimited.

We use the Lagrange multiplier method to find critical points $\vect{y}$ that are candidates for the optimal free responses. 
For the moment, we ignore the weighted decorrelation constraint (\ref{eq:wdec}) to solve for the first optimal free response, but we consider the remaining responses later. 
Due to the close relationship between GSFA and LPP, the approach below is strongly related to Laplacian Eigenmaps~\citep{Belkin-Niyogi-NC-2003}. 
Let
\begin{equation}
\label{eq:lagranian}
\Lag \eqdef \big( 2 - \frac{2}{R} \vect{y}^T \GG \vect{y} \big) + \alpha \vect{v}^T \vect{y} + \beta \big( \vect{y}^T \Diag(\vect{v}) \vect{y} - Q \big)
\end{equation}
be a Lagrangian corresponding to the objective function~(\ref{eq:wobj}), under the constraints (\ref{eq:wzm}) and (\ref{eq:wuv}). 
A signal $\vect{y}$ is a critical point if the partial derivatives of $\Lag$ with respect to $\alpha, \beta$, and $y(n)$, for $1 \le n \le N$, are simultaneously zero:

\begin{align}
\label{eq:partialAlpha}
\partial \Lag / \partial \alpha \stackrel{(\ref{eq:lagranian})}{=} \vect{v}^T \vect{y} &\eqlet 0 \, ,  \\
\label{eq:partialBeta}
\partial \Lag / \partial \beta \stackrel{(\ref{eq:lagranian})}{=} \vect{y}^T \Diag(\vect{v}) \vect{y} - Q &\eqlet 0 \, \ttxt{, and} \\
\label{eq:partialY}
\partial \Lag / \partial \vect{y} \stackrel{(\ref{eq:lagranian})}{=} - \frac{4}{R} \GG \vect{y} + \alpha \vect{v} + 2 \beta \Diag(\vect{v}) \vect{y} &\eqlet \vect{0} \, ,
\end{align}
where $\vect{0}$ is a vector of zeros.

Equations (\ref{eq:partialAlpha}) and (\ref{eq:partialBeta}) merely require that the output $\vect{y}$ has weighted zero mean and weighted unit variance, respectively.
Multiplying (\ref{eq:partialY}) with $\vect{1}^T$ from the left and taking into account that $\vect{1}^T \Diag(\vect{v}) = \vect{v}^T$, $\vect{1}^T \vect{v} \stackrel{(\ref{eq:Q})}{=} Q$,$\vect{1}^T \GG \stackrel{(\ref{eq:consistency})}{=} \frac{R}{Q} \vect{v}^T$, and $Q>0$ results in:
\begin{equation}
\label{eq:valueAlpha2}
- \frac{4}{R} \Big( \frac{R}{Q} \vect{v}^T \Big) \vect{y}  + \alpha Q + 2 \beta \vect{v}^T \vect{y} = 0 \, ,
\end{equation}
implying  $\alpha=0$ due to (\ref{eq:partialAlpha}). Therefore, (\ref{eq:partialY}) can be simplified to:
\begin{align}
\label{eq:partialY_b}
&& \Big(-\frac{4}{R} \GG + 2 \beta \Diag(\vect{v}) \Big) \vect{y} &= \vect{0} \, , \\
\label{eq:partialY_c}
&\iif& \quad \Diag(\vect{v}^{-1/2}) \Big(\frac{4}{R} \GG - 2 \beta \Diag(\vect{v}) \Big) \Diag(\vect{v}^{-1/2}) \Diag(\vect{v}^{1/2}) \vect{y} &= \vect{0} \, , \\
\label{eq:partialY_d}
&\iif& \quad \Big( \frac{4}{R}\Diag(\vect{v}^{-1/2}) \GG \Diag(\vect{v}^{-1/2}) - 2 \beta \vect{I} \Big) \big( \Diag(\vect{v}^{1/2}) \vect{y} \big) &= \vect{0} \, , \\
\label{eq:partialY_e}
&\iif& \quad  \Big( \Diag(\vect{v}^{-1/2}) \GG \Diag(\vect{v}^{-1/2}) -  \frac{R \beta}{2} \vect{I} \Big) \big( \Diag(\vect{v}^{1/2}) \vect{y} \big) &= \vect{0} \, ,
\end{align}
where $\vect{v}^{1/2}$ is defined as the element-wise square root of the elements of $\vect{v}$, and $\vect{v}^{-1/2}$ is defined similarly (as usual, weights $v_j$ are required to be strictly positive).

In a few words, $\vect{y}$ is a critical point if it fulfills the weighted normalization conditions and the vector $\Diag(\vect{v}^{1/2}) \vect{y}$ is an eigenvector of the matrix $\MM$ defined as 
\begin{equation}
\label{eq:M_in_Gamma}
\MM \eqdef \Diag(\vect{v}^{-1/2}) \GG \Diag(\vect{v}^{-1/2}) \, .
\end{equation}
The corresponding eigenvalue is denoted 
\begin{equation}
\label{eq:lambda_in_beta}
\lambda = \frac{R \beta}{2} \, . 
\end{equation}
We denote the (orthogonal) eigenvectors of the matrix $\MM$ as $\vect{u}_j$ with $\vect{u}_j^T\vect{u}_j=1$. Each eigenvector gives rise to a critical point $\vect{y}_j \eqdef Q^{1/2}\Diag(\vect{v}^{-1/2}) \vect{u}_j$ if the weighted normalization conditions are also satisfied by $\vect{y}_j$.
The slowest possible solution is the critical point $\vect{y}_j$ 
with the smallest $\Delta$-value.
As we show below, the $\Delta$-value of a critical point $\vect{y}_j$ is directly related to the eigenvalue $\lambda_j$ of the eigenvector $\vect{u}_j = Q^{-1/2}\Diag(\vect{v}^{1/2}) \vect{y}_j$ of $\MM$ and can be computed as follows.

\begin{align}
\Delta_{\vect{y}_j} &\stackrel{(\ref{eq:wobj})}{=}  2 - \frac{2}{R} (\vect{y}_j)^T \GG \vect{y}_j \\
&\stackrel{(\ref{eq:M_in_Gamma})}{=} 2 - \frac{2}{R} (\vect{y}_j)^T \Diag(\vect{v}^{1/2}) \MM \big( \Diag(\vect{v}^{1/2}) \vect{y}_j \big) \\
&\stackrel{(\ref{eq:lambda_in_beta})}{=} 2 - \frac{2}{R} (\vect{y}_j)^T \Diag(\vect{v}^{1/2}) \lambda_j \Diag(\vect{v}^{1/2}) \vect{y}_j \\
\label{eq:delta_in_lambda}
&\stackrel{(\ref{eq:wuv})}{=} 2 - \frac{2Q}{R} \lambda_j
\end{align}

Thus, the slowest solution is the critical point $\vect{y}_j$ with the largest eigenvalue $\lambda_j$. 
The remaining optimal free responses are the remaining critical points, and their 
the corresponding eigenvalue defines their order, from largest to smallest. The weighted decorrelation condition (\ref{eq:wdec}) is fulfilled due to the orthogonality of the eigenvectors: $\vect{u}_i^T \vect{u}_j = 0 \Leftrightarrow \frac{1}{Q} \vect{y}_i^T \Diag(v)\vect{y}_j=0$ (follows from the definition of $\vect{y}_j$ above).

One special case is when an eigenvalue has multiplicities. This means that two or more optimal free responses and any rotation of them have the same delta value. In this case, some optimal free responses are not unique, but any rotation of them is equivalent.

\subsection{Design of a Training Graph for Learning One or Multiple Labels}
\label{sec:ELL}
Given a set of samples $\{ \vect{x}(1), \dots, \vect{x}(N) \}$ with label $\Bell=(\ell_1, \dots, \ell_N)$, we show how to generate a training graph, such that the slowest feature that could be extracted by GSFA is equal to a normalized version of the label. 
Notice that this problem (determining the structure of a training graph, or more concretely, its edge-weight matrix $\GG$, having a particular optimal solution)  differs considerably from the original GSFA problem of finding an optimal solution 
given a training graph and a feature space. 
The approach can be extended to multiple labels per sample. To distinguish them, we introduce an index $1 \le j \le L$, making $\Bell_j$ denote the $j$-th label. In this case, the $L$ labels can be expressed linearly in terms of the first $L$ free responses.

Vertex-weights $v_n$ indicate {\it a priori} likelihood information about the samples, and are thus assumed to be given and strictly positive. If this information is absent, one may set the vertex weights constant, e.g.\ $\vect{v} = \frac{1}{N}\vect{1}$.

Due to the normalization constraints, the outputs generated by GSFA must have weighted zero mean (\ref{eq:wzm}) and weighted unit variance (\ref{eq:wuv}).
Therefore, we learn a weight-normalized label $\tilde{\Bell}$, 
as follows:
Let $\mu_{\ell} = \frac{1}{Q} \vect{v}^T \Bell$ be the weighted label average 
and $\sigma^2_{\Bell} = \frac{1}{Q} (\Bell-\mu_{\ell}\vect{1}) ^ T \Diag(\vect{v}) (\Bell-\mu_{\ell}\vect{1})$ be the weighted label variance.
Then, the normalized label is computed as 
\begin{align}
\label{eq:label_normalization}
\tilde{\Bell} = \frac{1}{\sigma_{\Bell}} (\Bell - \mu_{\ell}\vect{1}) \, .
\end{align}
Hence, it is trivial to map from a normalized to a non-normalized label and {\it vice versa}. 

In order for the construction to work when samples have multiple labels, we must weight decorrelate them first. 
To decorrelate two labels $\Bell_{j'}$ and $\Bell_{j}$, with $j'>j$, one can project $\Bell_{j}$ out of $\Bell_{j'}$;
$\ell_{j'}^\ttxt{dec}(n) = \ell_{j'}(n) - \frac{1}{Q} \big( \Bell_{j'}^T \Diag(\vect{v}) \Bell_{j} \big) \ell_{j}(n)$, which is an invertible linear operation.

From now on, we assume that the labels $\Bell_1, \dots, \Bell_L$ are decorrelated and normalized.
We compute edge weights $\e_{n,n'}$ such that the $j$-th optimal free response
is equal to $\Bell_j$ (with arbitrary polarity).

Define
\begin{equation}
\label{eq:GGELL}
\GGELL \eqdef \Diag(\vect{v}^{1/2}) \MMELL \Diag(\vect{v}^{1/2}) \, ,
\end{equation}
where
\begin{equation}
\label{eq:MMELL}
\MMELL \eqdef \sum_{j=0}^{N-1} \lambda_j \vect{u}^\ttxt{\tiny ELL}_j (\vect{u}^\ttxt{\tiny ELL}_j)^T\, .
\end{equation}

If $L<N-1$ one can set $\lambda_{j>L}=0$. The matrix $\GGELL$ defined above is symmetric by construction.
The eigenvectors and eigenvalues of $\MMELL$, which are explicit in its eigenvector decomposition above, directly define the matrix $\GGELL$ and determine the optimal free responses of the resulting graph.
Concretely, for each $j \ge 1$ one sets $\vect{u}^\ttxt{\tiny ELL}_j$ according to the desired label $\Bell_j$ (ignore $\vect{u}^\ttxt{\tiny ELL}_0$ and $\lambda_0$ for the time being).
\begin{align}
\label{eq:u_in_ell}
\vect{u}^\ELL_j = Q^{-1/2} \Diag(\vect{v}^{1/2}) \Bell_j, \ttxt{ for } j \ge 1 
\end{align}

Notice that the weighted decorrelation of the labels translates directly into the orthogonality of the corresponding eigenvectors
\begin{align}
\label{eq:orthogonality_u}
\frac{1}{Q}(\Bell_j)^T \Diag(\vect{v}) \Bell_{j'} \stackrel{(\ref{eq:wdec})}{=} 0 \quad \stackrel{(\ref{eq:u_in_ell})}{\Leftrightarrow} \quad (\vect{u}^\ttxt{\tiny ELL}_j)^T\vect{u}^\ttxt{\tiny ELL}_{j'}=0
\end{align}

Once the eigenvectors are computed we must decide which eigenvalues we want to give them. Alternatively, we can decide which $\Delta$ values we give to the labels, because $\Delta_{\Bell_j}$ and $\lambda_j$ are directly related: $\lambda_j \stackrel{(\ref{eq:delta_in_lambda})}{=} \frac{R}{2Q} ( 2 - \Delta_{\Bell_j})$.

Larger eigenvalues (equivalent to smaller $\Delta$ values) might result in higher accuracy for the corresponding label. We give some intuition on how to choose the eigenvalues of the eigenvectors.
a) In general, important labels should have larger eigenvalues than less important ones.
b) The global scale of the eigenvalues $\lambda_{j>0}$ is irrelevant, only their relative scales matter. For convenience one can scale them so that $\sum \lambda_{j>0}=1$. 
c) If two labels are similarly important, their eigenvalues should be also similar.

For example, if one only wants to learn a single label $\Bell_1$ with a delta value $\Delta_{\Bell_1}=0$, one can set $\vect{u}^\ELL_1= Q^{-1/2}\Diag(\vect{v}^{1/2}) \Bell_1$, $\lambda_1=1$, and the eigenvalues $\lambda_{j>1}$ to zero.
If $\Bell_1$ takes only two possible values (e.g., $-1$ and 1), the resulting graph will be disconnected and contain two clusters. Otherwise, the resulting graph will be connected, and the condition $\Delta_{\Bell_1}=0$ necessarily implies that some of the resulting edge weights will be negative, a condition that we deal with in Section~\ref{sec:ConversionNegPosWeights}.



The analysis of Section~\ref{sec:critical_points}, which is used by the ELL method requires that the graph fulfills the consistency restriction (\ref{eq:consistency}), which depends on $\vect{u}_0$ and $\lambda_0$, as follows,
\begin{align}
\GGELL \vect{1} \;\;\;\; &\stackrel{\mathclap{(\ref{eq:GGELL},\ref{eq:MMELL})}}{=} \;\;\;\; \Diag(\vect{v}^{1/2}) \big( \sum \lambda_j  \vect{u}^\ttxt{\tiny ELL}_j (\vect{u}^\ttxt{\tiny ELL}_j)^T \big) \Diag(\vect{v}^{1/2}) \vect{1} \\
&\stackrel{}{=} \;\; \Diag(\vect{v}^{1/2})  \big( \sum \lambda_j  \vect{u}^\ttxt{\tiny ELL}_j (\vect{u}_j^\ttxt{\tiny ELL})^T \big) \vect{u}_0^\ttxt{\tiny ELL} Q^{1/2} \\
&\stackrel{}{=} \;\; \Diag(\vect{v}^{1/2}) \lambda_0 \vect{u}_0^\ttxt{\tiny ELL} Q^{1/2} \\
&\stackrel{!}{=} \;\; (R/Q) \vect{v} \, .
\end{align}

To satisfy the equation above one can set $\vect{u}^\ELL_0 = Q^{-1/2}\vect{v}^{1/2}$ with eigenvalue $\lambda_0 = R/Q$, which also ensures $(\vect{u}^\ELL_0)^T \vect{u}^\ELL_0 = 1$ and $\vect{1}^T \GGELL \vect{1} = R$. 
The free pseudo-response $\Bell_0 \stackrel{(\ref{eq:u_in_ell})}{=} \vect{1}$ corresponding to $\vect{u}_0^\ttxt{\tiny ELL}$ fulfills equations (\ref{eq:wuv}) and (\ref{eq:wdec}) but not (\ref{eq:wzm}). Therefore, $\Bell_0$ is not a feasible solution, but it has similar properties to the optimal free responses.
The introduction of $\vect{u}^\ttxt{\tiny ELL}_0$ does not reduce the generality of the labels $\Bell_{j>0}$ that can be learned; orthogonality between  $\vect{u}^\ttxt{\tiny ELL}_0$ and $\vect{u}^\ttxt{\tiny ELL}_{j>0}$ is equivalent to the weighted zero mean of $\Bell_{j>0}$ (\ref{eq:wzm}), which is required anyway for any feasible solution, i.e., $(\vect{u}^\ttxt{\tiny ELL}_0)^T\vect{u}^\ttxt{\tiny ELL}_j=0 \stackrel{(\ref{eq:orthogonality_u})}{\iif} (Q^{-1/2}\vect{v}^{1/2})^T Q^{-1/2} \Diag(\vect{v}^{1/2}) \Bell_j= Q^{-1}\vect{v}^T \Bell_j = 0$.

Although only $L$ free responses are explicitly defined, $N-L-1$ additional optimal free responses are defined implicitly with an eigenvalue of 0, corresponding to $\Delta=2.0$. This $\Delta$ value has a particular meaning, because as we prove in the next paragraph, it is the $\Delta$ value of unit-variance zero-mean i.i.d.\ noise for certain graphs. 




\subsubsection{Expected Weighted $\Delta$ Value of a Noise Feature}
Let $\vect{y}$ be a noise feature randomly sampled from a zero-mean unit-variance distribution $\DDist$, i.e., $y(n) \gets \DDist(0,1)$.
On average, $\vect{y}$  fulfills the normalization conditions, as can be seen as follows. 
\begin{align}
&\ttxt{(\ref{eq:wzm}):}& \;\;\; \langle \vect{v}^T \vect{y} \rangle_\DDist \; &= \; \vect{v}^T \langle \vect{y} \rangle_\DDist = 0 \, , \\ 
&\ttxt{(\ref{eq:wuv}):}& \;\;\; \langle \vect{y}^T \Diag(\vect{v}) \vect{y} \rangle_\DDist \; &= \; \langle \sum_n v_n y(n)^2 \rangle_\DDist =  \sum_n v_n \langle y(n)^2 \rangle_\DDist = Q \, ,
\end{align} 
where $\langle \cdot \rangle_\DDist$ denotes expected value when sampling over $\DDist$. The expected delta value can be computed as
\begin{align}
&& \langle \Delta_{\vect{y}} \rangle_\DDist \; &\stackrel{\mathclap{(\ref{eq:wobj_long})}}{=} \; 
\frac{1}{R} \sum_{n,n'} \e_{n,n'} \langle (y(n')-y(n))^2 \rangle_\DDist \\
&& \; &= \; \frac{1}{R} \Big( \sum_{n,n',n\neq n'}  \e_{n,n'} \langle (y(n')-y(n))^2 \rangle_\DDist + \sum_{n}  \e_{n,n} \langle (y(n)-y(n))^2 \rangle_\DDist \Big) \\
&& \; &= \frac{1}{R} \Big( \sum_{n,n',n\neq n'}  \e_{n,n'} \big( \langle y(n')^2 \rangle_\DDist + \langle y(n)^2 \rangle_\DDist - 2 \langle y(n') \rangle_\DDist \langle y(n) \rangle_\DDist \big) + 0 \Big) \\
&& \; &= \; \frac{1}{R} \sum_{n,n',n\neq n'}  \e_{n,n'} (1 + 1 - 0) \\
&& \; &= \; \frac{2}{R} \big( \sum_{n,n'} \e_{n,n'} - \sum_{n} \e_{n,n} \big) = \frac{2(R-\sum_{n} \e_{n,n})}{R} \, .
\end{align} 

Therefore, if the graph has no self-loops (i.e., $\forall n:\e_{n,n}=0$), the expected $\Delta$ value $\langle \Delta_{\vect{y}} \rangle_\DDist$ of a noise feature is 2. The self-loops of a graph (e.g., one constructed using the ELL method) can be removed without changing the free responses, only affecting the scale of the delta values due to the change in $R$. The consistency restriction might be broken, though.

\ignore{
Actually only the feasible solution $\tilde{\Bell}_1$ has a small delta value.
Therefore, a surprising property of the neighborhood structure of the training graph defined with the eigenvalues/eigenvectors above, is that all features decorrelated with $\tilde{\Bell}_1$ are fast. For instance, weight-normalized versions of $(\tilde{\Bell}_1)^2, (\tilde{\Bell}_1)^3, \dots$ decorrelated with $\tilde{\Bell}_1$ would still have large $\Delta$-values.
Let us exemplify this. Assume that $\vect{v}=\vect{1}$, and $\tilde{\Bell}_1(n) = b_1 \cos(\frac{\pi  i}{N} )$, for $1 \le i \le N$, is the label that we want to learn, for a suitable constant $b_1$, and we use the procedure above to generate the graph. Functions such as $y_2(n) = b_2 \cos(\frac{2 \pi  i}{N} )$ or $y_3(n) = b_3 \cos(\frac{3 \pi  i}{N} )$ will have a delta value of 2, just as normalized noise does.
This counterintuitive result shows that due to the topology defined by the graph structure the function $b_1 \cos(\frac{\pi  i}{N} )$ is slow, while $b_2 \cos(\frac{2 \pi  i}{N} )$ and related harmonics are fast signals.
}

\subsection{Elimination of Negative Edge Weights}
\label{sec:ConversionNegPosWeights} 
From the objective function (\ref{eq:wobj_long}), it is obvious that a positive edge weight connecting two samples expresses that those samples should be mapped close to each other in feature space.
In contrast, negative edge weights express that two samples should be mapped as far apart as possible, thus encoding output dissimilarities. Nevertheless, the weighted unit variance constraint still applies, so the solutions are not unbounded.

If the edge weights are non-negative, the smallest possible $\Delta$ value  is $\Delta = 0$.
However, if negative edge weights are allowed, some feasible features might have $\Delta < 0$.
A feature with $\Delta<0$ would appear to be ``slower'' than the infeasible constant feature $\vect{y}=\vect{1}$ with $\Delta=0$, contradicting the intuitive interpretation of slowness.
Moreover, negative edge weights hinder the probabilistic interpretation of the graph (see Section~\ref{sec:LGSFA}), because some of the transition probabilities $\e_{n,n'}/R$ of the resulting Markov chain would be negative.

Training graphs constructed using the ELL method might include negative edge weights, which would result in the disadvantages described above. 

In this section, we add an additional step to the ELL method to ensure that the training graph has non-negative edge weights.
More concretely, we show how to transform a training graph with strictly positive vertex weights $v_n$ and arbitrary edge weights $\GG$ (positive and negative) into a graph with the same vertex weights and only non-negative edge weights $\GG'$. 
The optimization problem defined by $\GG'$ is equivalent to the original optimization problem in terms of its solutions and their order. Only the value of the objective function is linearly changed (or, more precisely, changed by an affine function).

Assume that $\forall n: v_n > 0$, and that there is at least one element $\e_{n,n'}<0$. Let $c \eqdef \ttxt{max}_{n,n'} \frac{-\e_{n,n'}}{v_n v_{n'}}$. The new edge weights $\GG'$ are defined as  
\begin{align}
\label{eq:non_negative_weights}
\GG' \eqdef \frac{1}{1+cQ^2/R}( \GG + c \vect{v}\vect{v}^T ).
\end{align} 

%

Now, we show the properties of $\GG'$ and its relation to $\GG$:
\begin{enumerate}
\item All elements of $\GG'$ are greater or equal to zero, as desired. (Follows from the definition of $c$; $ \e_{n,n'} + c v_n v_{n'} \ge 0$.)
\item Symmetry is preserved; clearly $\GG'$ is symmetric if and only if $\GG$ is symmetric.
\item The sum of edge-weights is preserved: 
\begin{align}
\label{eq:Rp_is_R}
R' = \vect{1}^T \GG' \vect{1} =  \frac{R+cQ^2}{1+cQ^2/R} = R \, .
\end{align}
\item Fulfillment of the graph consistency restriction (\ref{eq:consistency}) is preserved: 
\begin{align}
\vect{1}^T \GG  &\stackrel{(\ref{eq:consistency})}{=} R/Q \vect{v}^T &\Rightarrow&& \vect{1}^T \GG' \;\; &\stackrel{\mathclap{(\ref{eq:non_negative_weights})}}{=} \;\; \frac{1}{1+cQ^2/R}(\vect{1}^T \GG + c \vect{1}^T \vect{v}\vect{v}^T ) \\
&&&&&\stackrel{\mathclap{(\ref{eq:consistency})}}{=} \;\; \frac{1}{1+cQ^2/R}(R/Q \vect{v}^T + c Q \vect{v}^T ) \\
&&&&&= \;\; \frac{R/Q}{1+cQ^2/R} (1 + c Q^2/R) \vect{v}^T \\
&&&&&= \;\; R/Q \vect{v}^T \, .
\end{align}
\item $\GG$ and $\GG'$ define equivalent optimization problems. Let $\vect{y}$ be a feasible solution. The constraints of the optimization problem are independent of $\GG'$, and only the objective function is modified as follows:
\begin{align}
\Delta'_\vect{y} \;\; &\stackrel{\mathclap{(\ref{eq:wobj_simp})}}{=} \;\; 2 - \frac{2}{R'} \vect{y}^T \GG' \vect{y} \\ 
&\stackrel{\mathclap{(\ref{eq:non_negative_weights},\ref{eq:Rp_is_R})}}{=} \;\;\;\; 2 - \frac{2}{R (1+cQ^2/R)} \big( \vect{y}^T \GG \vect{y} + c \vect{y}^T \vect{v}  \vect{v}^T  \vect{y} \big) \\
&\stackrel{\mathclap{(\ref{eq:wzm})}}{=} \;\; 2 - \frac{2}{R (1+cQ^2/R)} \vect{y}^T \GG \vect{y} \\
\label{eq:delta_tmp}
&\stackrel{\mathclap{(\ref{eq:wobj_simp})}}{=} \;\; 2 - \frac{2}{R (1+cQ^2/R)} \frac{R}{2}  (2 - \Delta_\vect{y}) \\
&= \;\; \frac{1}{(1+cQ^2/R)} \big( \Delta_\vect{y}+ \frac{2 c Q^2}{R}  \big) \, . 
\end{align}
Therefore, the objective function is only modified by a positive scaling factor 
and a constant positive offset, 
proving that the optimal free solutions to the training graph remain stable, as well as their order.
\item In particular, a feature $\vect{y}$ with $\Delta_\vect{y}=2$ preserves its delta value, i.e.\ $\Delta_\vect{y} = 2 \stackrel{(\ref{eq:delta_tmp})}{\Leftrightarrow} \Delta_\vect{y}' = 2$.
\end{enumerate}

\subsection{Auxiliary Labels for Boosting Estimation Accuracy} 
\label{sec:auxiliary_labels}
When GSFA is applied repeatedly (e.g., cascaded or in a convergent hierarchical GSFA network) one can provide additional \emph{auxiliary} labels derived from the original one $\Bell_1$ to improve the estimation accuracy.
Informally, even though a given GSFA node might not be able to extract $\Bell_1$ accurately, it might be capable of approximating features $f_2(\Bell_1), f_3(\Bell_1), \dots, f_K(\Bell_1)$. Since these features are derived from the label, they contain information that (partially) determines it. Hence, a posterior GSFA node might be able to disentangle these features more effectively to recover the original label. 
One can explicitly promote the appearance of these features by learning also auxiliary labels $\Bell_k = f_k(\Bell_1)$, for $2 \le k \le K$. 

The functions $f_k$ can be defined arbitrarily, one simple choice is to use
\begin{align}
\label{eq:auxiliary_labels}
\ell_k(n) = \cos\Big(\frac{ \ell_1(n)-\min(\Bell_1)}{\max(\Bell_1) - \min(\Bell_1)} \pi k \Big) \, \ttxt{, for } 2 \le k \le K \, ,
\end{align}
where $\max(\Bell_1)$ is the largest label value, and $\min(\Bell_1)$ is the smallest one.
Notice that for $k=2$ the argument of the cosine function ranges from 0 to $\pi$, for $k=3$ from 0 to $(3/2)\pi$, etc. In this sense, these features are ``higher-frequency'' versions of $\Bell_1$.
The eigenvalues corresponding to the auxiliary labels must be set smaller than those of the original label. Otherwise, the slowest features found might be close to the auxiliary labels rather than to the original one. 
From now on, we use the term \emph{target} labels to refer to the original and auxiliary labels, if present.

Interestingly, in regular SFA (or GSFA trained with the reordering graph) the inclusion of auxiliary labels occurs automatically. The slowest free response is a half period of a cosine function, and the subsequent free responses are the higher-frequency harmonics of the first one (see Section~\ref{sec:analysis_training_graphs}, particularly Figure~\ref{fig:graph_analysis}).

\subsection{Computational Complexity of Explicit Label Learning}
The main drawback of ELL is its computational efficiency compared to efficient pre-defined training graphs, which is more marked for large $N$. We analyse the efficiency of explicit label learning by considering its two main parts: The construction of the training graph and training GSFA with it.

The graph construction requires $\bigO(L^2 N + L N^2)$ operations. The term $L^2 N$ is due to the transformation of $L$ target labels into eigenvectors, which might require a decorrelation step on $L$ $N$-dimensional vectors.
The term $LN^2$ is due to the computation of $\MM$, which involves $L$ vector multiplications $\vect{u}_j \vect{u}_j^T$. 

When training GSFA, three computations are particularly expensive.
Firstly, the computation of $\Cov_{\vect{G}}$, which takes $\bigO(N I^2)$ operations.
Secondly, the computation of $\DCov_{\vect{G}}$, which can be expressed as
$\DCov_{\vect{G}} = \frac{2}{Q} \vect{X}\Diag(\vect{v})\vect{X}^T- \frac{2}{R} \vect{X} \GG \vect{X}^T$, where $\vect{X}=\big(\vect{x}_1, \dots, \vect{x}_N\big)$, taking $\bigO(N^2 I + N I^2)$ operations.
Thirdly, the solution to the generalized eigenvalue problem, which requires $\bigO(I^3)$ operations. 
Therefore, in general, training GSFA requires $\bigO(NI^2 + N^2I + I^3)$ operations.  Typically $N > I$ to avoid overfitting, so the computation of $\DCov_{\vect{G}}$ is the most expensive part.

However, when a pre-defined graph is used instead of an ELL graph, $\DCov_{\vect{G}}$ might be computed more efficiently by using optimized algorithms. 
If an efficient pre-defined graph is used (e.g.\ the serial graph), $\DCov_{\vect{G}}$ can be computed in $\bigO(N I^2)$ operations (due to the regular structure of the graph),
which is the complexity of the same operation using standard SFA on $N$ $I$-dimensional samples. 
Moreover, if the number of edges $N_e \le N (N+1)/2$ is small, one can use (\ref{eq:dcov_tg}) to compute $\DCov_{\vect{G}}$ in $\bigO(N_e I^2)$ operations.
Therefore, for these two special cases, training GSFA takes $\bigO(NI^2 + I^3)$ and $\bigO((N_e+N)I^2 + I^3)$ operations, respectively.

\section{Applications of Explicit Label Learning}
\label{sec:applications_ELL}
In this section, we present three applications of the proposed method. First, we illustrate how  to solve a regression problem with GSFA explicitly, learning a direct mapping from images to their labels (see Figure~\ref{fig:ELLidea}.c). 
In the second application, we analyse two pre-defined graphs by computing their optimal free responses.
In the third application, we use the ELL method in a new way to learn compact discriminative labels for classification.

\subsection{Explicit Estimation of Gender with GSFA}
\label{sec:gender_estimatin_ELL}
We consider the problem of gender estimation from artificial face images, which is treated here as a regression problem, because the gender parameter is defined as a real value by the face modelling software. 

\begin{figure}[hbt!]
\begin{small}
\begin{center}
\includegraphics[width=0.85\textwidth]{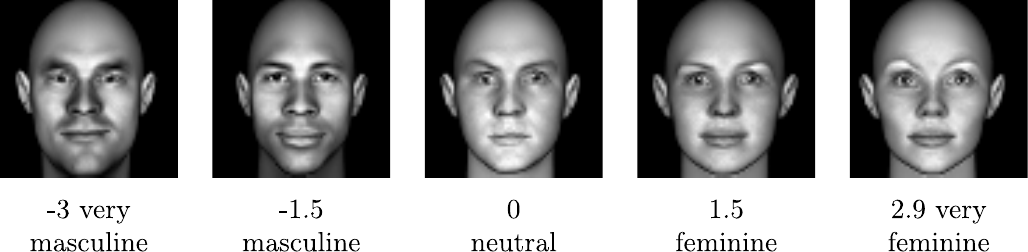}
\caption{Example of the images used, showing different values of the gender parameter.}
\label{fig:examples_gender}
\end{center}
\end{small}
\end{figure}

\paragraph{Input data.}
The input data are 20,000 64$\times$64 grayscale images. Each image is generated using a new subject identity, where the gender is specified by us, and the rest of the parameters of the faces (e.g., age, racial composition) are random. 
The average pixel intensity of each image is normalized by multiplying it by an appropriate factor. 
The resulting images show subjects with a fixed pose, no hair or accessories, fixed illumination, constant average pixel intensity, and a black background. See Figure~\ref{fig:examples_gender} for some sample images.
$60$ different values of the gender parameter are used ($-3, -2.9, \dots, 2.9$).

The images are randomly split into a training and a test set.
The training set consists of 10,800 images, $180$ images for each gender value, whereas the test set consists of 1,200 images, $20$ images for each gender value.

Besides the gender label, we also consider a second ``color'' label, which is the average pixel intensity of the image \emph{before} normalization.
Due to normalization, this label cannot be computed directly, but it can be estimated from other cues, such as the subject's apparent race and face size. In the following experiment we consider only the gender label, but later we use both labels (gender and color) simultaneously.



\paragraph{Network used.}
For efficiency reasons, hierarchical GSFA (HGSFA) is used. We teste an 8-layer HGSFA network with the structure described in Table~\ref{tab:network_architecture}.
The nodes of the network have non-overlapping receptive fields and are composed of an expansion function 
$(x_1, \dots, x_n) \mapsto (x_1, \dots, x_n, |x_1|^{0.8}, \dots, |x_n|^{0.8})$ followed by linear GSFA.
This expansion only doubles the data dimensionality and is called 0.8Expo~\citep{EscalanteWiskott-2011}. The nodes of the first layer include a PCA pre-processing step, in which 50 out of 64 components are preserved.

\begin{table}[hbt!]
\small
\begin{center}
\begin{tabular}{c|c|c|c|c|c}
\multirow{2}{*}{layer} & number      & node's receptive & input dim & expanded dim & output dim \\
                       & of nodes    & field (pixels) & per node  & per node & per node   \\
\hline
1     & 8$\times$8    & 8$\times$8   & 64 & 100 & 40 \\
2     & 4$\times$8    & 16$\times$8  & 80 & 160 & 40 \\
3     & 4$\times$4    & 16$\times$16 & 80 & 160 & 40 \\
4     & 2$\times$4    & 32$\times$16 & 80 & 160 & 40 \\
5     & 2$\times$2    & 32$\times$32 & 80 & 160 & 40 \\
6     & 1$\times$2    & 64$\times$32 & 80 & 160 & 40 \\
7     & 1$\times$1    & 64$\times$64 & 80 & 160 & 40 \\
8     & 1$\times$1    & 64$\times$64 & 40 &  80 &  6 \\
\end{tabular}
\caption{\label{tab:network_architecture}
Structure of the GSFA hierarchical network. 
The inputs to the nodes in the first layer are 8$\times$8-pixel patches. The input to the node in layer 8 is the output of the node in layer 7. The inputs to all other nodes come from two contiguous (either vertically or horizontally) nodes in the previous layer.}
\end{center}
\end{table}

\paragraph{Training graphs for gender estimation.} Several training graphs are constructed with the ELL method described in Sections~\ref{sec:ELL}--\ref{sec:auxiliary_labels}.
The graphs are denoted $\ttxt{ELL}^g$-$L$, where $L$ is the total number of target labels considered, with $L \in \{1,10,20,30,40, 50\}$. 
The first target label $\ell_1(n)$ is the gender parameter, where $1 \le n \le$ 10,800.
The remaining $L-1$ labels are auxiliary and computed using (\ref{eq:auxiliary_labels}).
For comparison purposes, the serial and reordering training graphs were also tested.

\paragraph{Label estimations.}
We used three mappings from the slowest features to the label estimation $\hat{\ell}$.
The first mapping (only available for the ELL graphs) is a linear scaling $\hat{\ell} = \pm y_1 \sigma_\ell+\mu_\ell$ that inverts the label normalization~(\ref{eq:label_normalization}). Since the sign of $y_1$ is arbitrary, it is globally adjusted to best fit the labels. 
The second method is linear regression (LR).
For these two methods, final label estimation $\hat{\ell}$ is clipped to the valid label range $[-3, 2.9]$.
The third mapping is the Soft GC method, which provides a soft estimation based on the class probabilities estimated by a Gaussian classifier \citep[trained with 60 classes,][]{EscalanteWiskott-2013b}.

\paragraph{Results.}
Table~\ref{tab:accuracy_ELL} (left) shows the label estimation errors (RMSE) for the gender label.
Depending on the mapping, the $\ttxt{ELL}^g$-10 and $\ttxt{ELL}^g$-40 graphs outperform the rest. This supports the intuition that auxiliary labels are useful.
$50$ target labels perform worse than $40$, probably in part because the output dimensionality of the intermediate nodes in the network is 40.
Without the final clipping step LR was clearly more accurate than linear scaling (experiment not shown), but both methods have similar accuracy if clipping is enabled.
For all graphs, the explicitly supervised soft GC method provided better accuracy than the linear scaling method, although the difference is less than one might have expected. 

\begin{table}[thb!]
\small
\begin{center}
\begin{tabular}{c|c|c|c|c}
\multicolumn{5}{c}{\bf Graph $\ttxt{ELL}^g$-$L$} \\
\multirow{2}{*}{$L$} & scaling   & LR          & soft GC & soft GC \\
                     & (1F)      & (1F)        & (1F)    & (3F)    \\                   
\hline
1                  & 0.376       & 0.380       & 0.364       & 0.365  \\
10                 & {\bf 0.364} & {\bf 0.365} & 0.353       & 0.356  \\
20                 & 0.372       & 0.374       & 0.356       & 0.357  \\
30                 & 0.367       & 0.368       & 0.350       & 0.349  \\
40                 & 0.368       & 0.367       & {\bf 0.346} & {\bf 0.345} \\
50                 & 0.376       & 0.375       & 0.351       &  0.350  \\
\end{tabular}
\hspace{1em}
\begin{tabular}{c|c|c|c|c}
\multicolumn{5}{c}{\bf Graph $\ttxt{ELL}^{g,c}$-$L$} \\
\multirow{2}{*}{$L$} & scaling   & LR        & soft GC & soft GC\\
                     & (1F)      & (1F)      & (1F)    & (3F)   \\                   
\hline
$2\times 1$        & {\bf 0.298} & {\bf 0.299}  & {\bf 0.289} & 0.284 \\
$2\times 5$        &  0.349      & 0.350        & 0.343       & {\bf 0.277}\\
$2\times 10$       &  0.423      & 0.426        & 0.410       & 0.288 \\
$2\times 15$       &  0.473      & 0.478        & 0.453       & 0.291 \\
$2\times 20$       &  0.508      & 0.514        & 0.479       & 0.292 \\
$2\times 25$       &  0.535      & 0.543        & 0.499       & 0.294 \\
\end{tabular}
\caption{\label{tab:accuracy_ELL}
Gender estimation errors (RMSE) using various graphs and either one (1F) or three (3F) features.
For the linear regression (LR) mapping, the label is estimated as $\hat{\ell}_1 = a y_1 + b$, with $a$ and $b$ fitted to the training data.
Chance level (RMSE) is 1.731 if one uses the constant estimation $\hat{\ell}_1 = -0.05$. All results on test data and averaged over 10 runs. 
(Left) Error using training graphs for gender estimation only. (Right) Error using training graphs for the experiment on simultaneous gender and color estimation.
}
\end{center} 
\end{table}

For comparison, the serial graph results in RMSEs of 0.351 (soft GC, 1F) and 0.349 (soft GC, 3F), whereas the reordering graph results in RMSEs of 0.353 (soft GC, 1F) and 0.347 (soft GC, 3F).
The accuracy of these two graphs appears to be similar; however, in more complex experiments the serial graph has typically been more accurate \citep[e.g.,][]{EscalanteWiskott-2013b}.
The $\ttxt{ELL}^g$-40 graph is, therefore, slightly more accurate than the serial and reordering graphs but 25 times slower, taking about 250 min for training instead of about 10 min (single thread).
		

\paragraph{Simultaneous learning of gender and color.}
We construct a graph that codes gender and color simultaneously, learning labels $\vect{y}_1,\dots,\vect{y}_{L}$, where $\vect{y}_1, \vect{y}_3, \dots, \vect{y}_{L-1}$ are derived from the gender label, and $\vect{y}_2, \vect{y}_4,\dots, \vect{y}_{L}$ are derived from the color label. These labels are computed similarly to when learning gender only but using two different original labels. We use linearly decreasing eigenvalues.
The resulting graphs are denoted $\ttxt{ELL}^{g,c}$-$L$, where $L$ is the total number of target labels, with $L=2 \times d$, and $2(d-1)$ is the number of auxiliary labels used for gender and color.

\begin{table}[hbt!]
\begin{center}
\small
\begin{tabular}{c|c|c|c|c}
\multicolumn{5}{c}{\bf Graph $\ttxt{ELL}^c$-$L$} \\
\multirow{2}{*}{$L$} & scaling   & LR          & soft GC     & soft GC \\
                     & (1F)      & (1F)        & (1F)        & (3F)    \\                   
\hline
1                  &   2         & 1.987       & 1.971       & 1.979 \\
10                 & {\bf 1.969} & {\bf 1.958} & 1.905       & 1.922 \\
20                 & 2.006       & 1.999       & 1.914       & 1.922 \\
30                 & 1.991       & 1.989       & 1.877       & 1.889 \\
40                 & 1.990       & 1.990       & {\bf 1.864} & {\bf 1.867} \\
50                 & 1.997       & 1.997       & 1.865       & 1.871 
\end{tabular}
\hspace{1em}
\begin{tabular}{c|c|c|c|c}
\multicolumn{5}{c}{\bf Graph $\ttxt{ELL}^{g,c}$-$L$} \\
\multirow{2}{*}{$L$} &  LR      & soft GC    &  LR          & soft GC   \\
                     & (1F)     & (1F)       & (3F)         & (3F)      \\                   
\hline
$2\times 1$          & 4.247    &  4.291     &  1.393       &  1.221    \\
$2\times 5$          & 3.606    &  3.614     &  {\bf  1.239}&  1.210    \\
$2\times 10$         & 3.214    &  3.185     &  1.337       &  1.180    \\
$2\times 15$         & 2.978    &  2.945     &  1.429       &  1.158    \\
$2\times 20$         & 2.828    &  2.802     &  1.501       &  1.141    \\
$2\times 25$         &{\bf 2.718}&{\bf 2.700}&  1.582       & {\bf 1.140}    
\end{tabular}
\caption{\label{tab:accuracy_ELL_color}
\emph{Color} estimation errors (RMSE) using various graphs and either one (1F) or three (3F) features.
Chance level (RMSE) is 7.447. All results on test data and averaged over 10 runs. 
(Left) Error using training graphs that code only color. (Right) Error using training graphs that simultaneously code gender and color. 
}
\end{center} 
\end{table}

The effect of coding gender and color simultaneously on gender estimation is shown in Table \ref{tab:accuracy_ELL}, right (compare to Table \ref{tab:accuracy_ELL}, left).
The $\ttxt{ELL}^{g,c}$-$L$ graphs yield higher accuracy than the $\ttxt{ELL}^g$-$L$ graphs.
The results on color estimation using the $\ttxt{ELL}^{g,c}$-$L$ graphs are shown in Table \ref{tab:accuracy_ELL_color}, right (compare to Table \ref{tab:accuracy_ELL_color}, left).
The slowest feature extracted represents mostly gender. However, it also contains color information because it allows color estimation better than the chance level.
When 3 features are preserved, the $\ttxt{ELL}^{g,c}$-$L$ graphs yield higher accuracy than the $\ttxt{ELL}^c$-$L$ graphs.
Similar experimental results have been reported, e.g.\ by \cite{Guo-Mu-IVC-2014}, who have shown that age estimation improves when gender and race labels are also considered.

\paragraph{Learning label transformations.} 
We verify that the method can learn other labels implicitly described by the data. More precisely, we use GSFA to learn labels $(\Bell_1)^2$ and $(\Bell_1)^3$, which are distorted versions of the original gender label $\Bell_1$.
The graphs constructed for this purpose are denoted $\ttxt{ELL}^g$-$40^{(\Bell_1)^2}$ and $\ttxt{ELL}^g$-$40^{(\Bell_1)^3}$, respectively.
Both of them include 39 auxiliary labels besides the main distorted label.
To better approximate the target labels, more complex nonlinearities are used in some of the nodes of the hierarchical networks.
The $(\Bell_1)^2$ network is identical to the  $\Bell_1$ network, except that in the top node the quadratic expansion is used instead of the 0.8Expo expansion.
Similarly, the $(\Bell_1)^3$ network uses the quadratic expansion in the 7th layer, and the 6th-degree polynomial expansion in the top node. In both networks, the output dimension of the node in the 7th layer is set to 3 to avoid overfitting due to the expansion in the 8th layer.

\begin{figure}[bht!]
\begin{small}
\begin{center}
\includegraphics[width=0.99\textwidth]{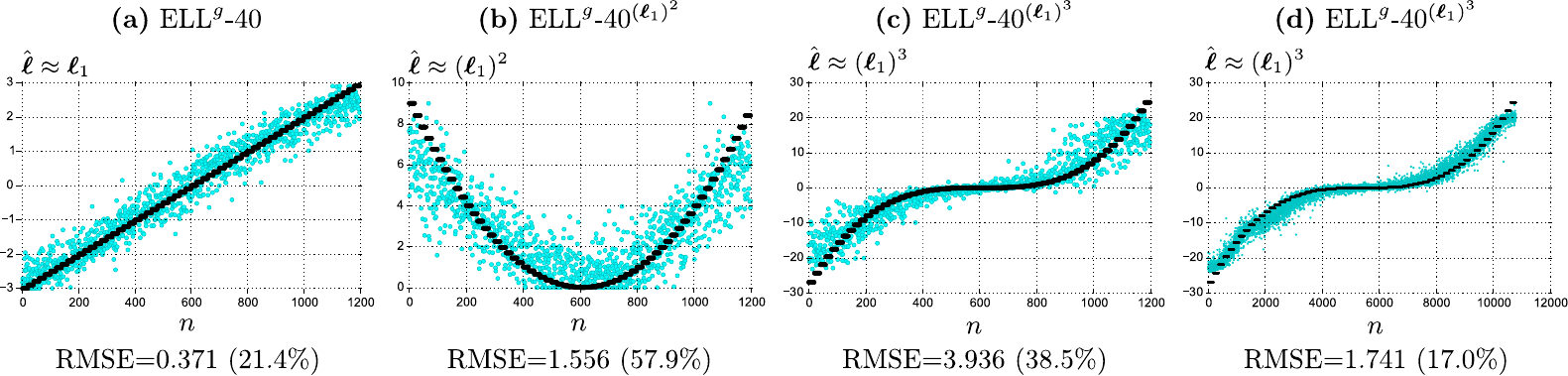}
\caption{
Plots (a) to (c) show the label estimations on test data when different distorted versions of $\Bell_1$ are learned. The linear scaling mapping was used. Therefore, the estimations are only generated from the slowest feature. Ground-truth values shown in thicker black. 
The RMSE is expressed in brackets as a percentage of the chance level. Plot (d) is analogous to (c) but shows training data.}
\label{fig:gender_outputs}
\end{center}
\end{small}
\end{figure}

The corresponding label estimations are shown in Figure~\ref{fig:gender_outputs}. For comparison, also the $\ttxt{ELL}^g$-$40$ graph is included.
The results prove that the ELL method can also be used to learn distortions of the main label. Admittedly, the accuracy of the estimations (normalized by the respective chance levels) decreases even though we increase the complexity of the feature space. 

\subsection{Analysis of Pre-Defined Training Graphs}
\label{sec:analysis_training_graphs}
In this section, we use the method of Section~\ref{sec:critical_points} to extract the optimal free responses of three graphs (reordering, serial and ELL-4).
The optimal free responses and their $\Delta$ values (alternatively, the eigenvectors $\vect{u}_j$ and eigenvalues $\lambda_j$) fully characterize the properties of a training graph, and provide another representation of it that might be more useful in some scenarios.

\begin{figure}[phbt!]
\begin{small}
\begin{center}
\includegraphics[width=\textwidth]{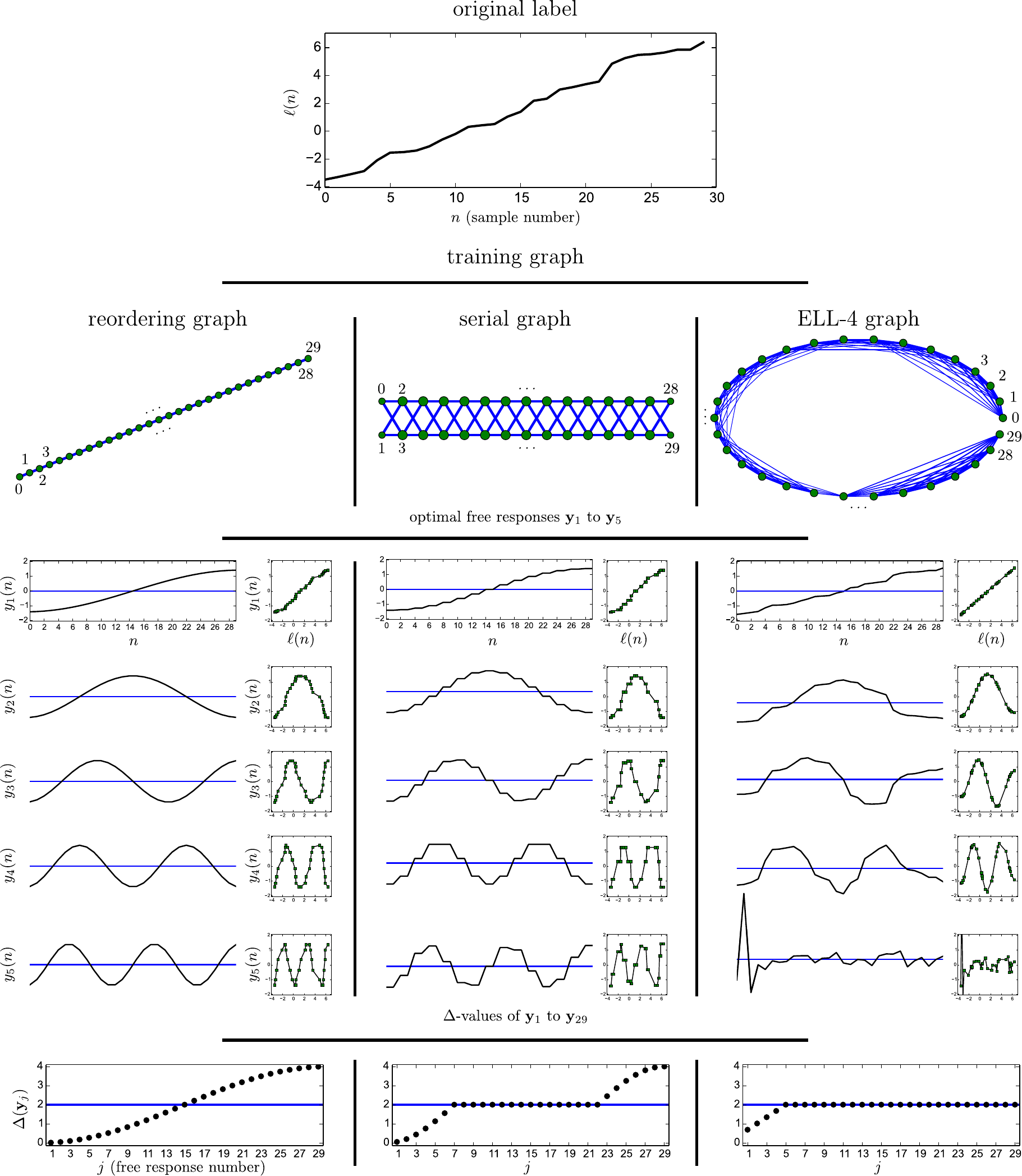}
\small
\caption{An arbitrary label (top) and three graphs that can be used to learn it. 
The five slowest optimal free responses of each graph are plotted, as well as the delta values of all optimal free responses. 
The ELL-4 graph is almost fully connected, but here only the strongest 30\% of the connections are displayed.
Samples have an index $n$ from 0 to 29, and free responses have an index $j$ from 1 to 29. 
The free responses are also plotted against the original label.
The polarity of the free responses was adjusted once to make them negative for the first sample.
}
\label{fig:graph_analysis}
\end{center}
\end{small}
\end{figure}

We compute optimal free responses using (\ref{eq:partialY_e})--(\ref{eq:lambda_in_beta}) and their delta values using (\ref{eq:delta_in_lambda}).
Therefore, these results have been obtained analytically. We plot them in Figure~\ref{fig:graph_analysis}, which shows an arbitrary label to be learned (top), and three different graphs that can be used for this purpose. 
Only $N=30$ samples (ordered by increasing label) were used to ease visualization, but the plots behave similarly for larger $N$.
The employed graphs are as follows. 
The reordering graph has been extended with two edge weights $\e_{0,0}=1$ and $\e_{N-1,N-1}=1$ to fulfill the consistency restriction (\ref{eq:consistency_long}), which is required by the method. These weights introduce a constant scaling $N/(N+2)$ of the delta values, without any further consequence.
The serial graph (Section~\ref{sec:clustered_serial}) has $K=15$ groups of 2 samples each.
The ELL-4 graph (Sections~\ref{sec:ELL}--\ref{sec:auxiliary_labels}) is constructed with the original labels $\ell_1(n)=\ell(n)$, and 3 auxiliary labels computed using (\ref{eq:auxiliary_labels}).


The figure shows that the most remarkable difference between the graphs is the number of optimal free responses with $\Delta < 2.0$, which is 14 for the reordering graph, 6 for the serial graph, and 4 for the ELL-4 graph, for the parameters above.
For arbitrary parameters, the reordering, serial and ELL-$L$ graphs have $\myfloor{(N-1)/2}$, $\myfloor{(K-1)/2}$, and, depending on the eigenvalues, up to $L \le N-1$ optimal free responses with $\Delta < 2.0$, respectively.

Although the graphs differ considerably in their connectivity, their first four to five optimal free responses have a somewhat similar shape. 
Since in all graphs the slowest free response $\vect{y_1}$ is increasing, a monotonic mapping would be enough to approximate the label for any of them.
However, for the serial graph the slowest response is constant within each group, which might lower accuracy due to a discretization error. 
The ELL-4 graph has been tailored to learn a particular label, and therefore $\vect{y_1}$ is exactly $\Bell_1$ (the original label) except for an offset and scaling.

The analysis makes clear that the serial and ELL-4 graphs are \emph{more selective} than the reordering graph regarding the features that they consider slow.
To illustrate why this might be an advantage, consider a 
scaled and noisy version $\hat{\vect{y}}_1$ of $\Bell_1$. More concretely, $\hat{y}_1(n) = \frac{\sqrt{2}}{2}\ell_1(n)+ \frac{\sqrt{2}}{2} e(n)$, where $e(n)$ is an i.i.d. zero-mean unit-variance noise signal.
When the reordering graph is used, the feature $\hat{\vect{y}_1}$ has an average $\Delta$-value of about 1 (i.e.\ $\langle \Delta_{\hat{\vect{y}}_1} \rangle \approx 1$), and therefore such a feature would appear to be faster than the auxiliary feature $\vect{y}_6$, because $\Delta_{\vect{y}_6} \approx 0.38$. Hence, a GSFA node trained with the reordering graph would favor the extraction of $\vect{y}_6$ over $\hat{y}_1(n)$, even though $\hat{y}_1(n)$ is more similar to the label. In contrast, the serial and ELL-4 graphs might favor the extraction of $\hat{\vect{y}}_1$, because for these graphs $\Delta_{\vect{y}_6}$ is close to 2.0.

\subsection{Compact Discriminative Features for Classification}
\label{sec:compact_features_ELL}
A well-known algorithm for supervised dimensionality reduction for classification is Fisher discriminant analysis (FDA). 
According to the theory of FDA, if there are $C$ classes, $C-1$ features define a $C-1$ dimensional subspace that best separates the classes.
In practice, one typically uses all these $C-1$ features, because all of them contain discriminative information and contribute to classification accuracy. 
The same holds for GSFA if the clustered training graph is used (GSFA+clustered), because in this case the features learned are equivalent to those of FDA~\citep{KlampflMaass:09b,EscalanteWiskott-2013b}. 

One can take advantage of hierarchical processing to do classification using the clustered graph (HGSFA+clustered). 
However, when the number of classes $C$ is large (e.g.\ $C \ge 100$) it might become expensive to preserve $C-1$ features in each node, because the size of the input to subsequent nodes would be a multiple of $C-1$. This dimensionality would be further increased by the expansion function, resulting in a large training complexity.
For instance, consider a 3-node nonlinear network for classification with two GSFA nodes in the first layer and one in the top. Suppose the first two nodes have output dimensionality $C-1=99$ so that the input into the top node would be 198, and suppose that the top node applies a quadratic expansion to its input data before linear GSFA.
The expanded data would have dimensionality $I'=$19,701.
The combination of a large sample dimensionality $I'$ and a large number of samples $N$ (with $N \gg I'$ to avoid overfitting) would result in considerable computational and memory costs.
Therefore, if we could code the class information in the first layer more compactly, we could reduce the output dimensionality of the first-layer nodes and reduce overfitting, aiming at increasing classification accuracy.

In this section, we use the theory of explicit learning with multiple labels to compute compact features for classification using GSFA.
We classify images of $C=32$ traffic signs from the German traffic sign recognition benchmark database \citep{Houben-IJCNN-2013}.

\begin{figure}[htb!]
\begin{small}
\begin{center}
\includegraphics[width=0.75\textwidth]{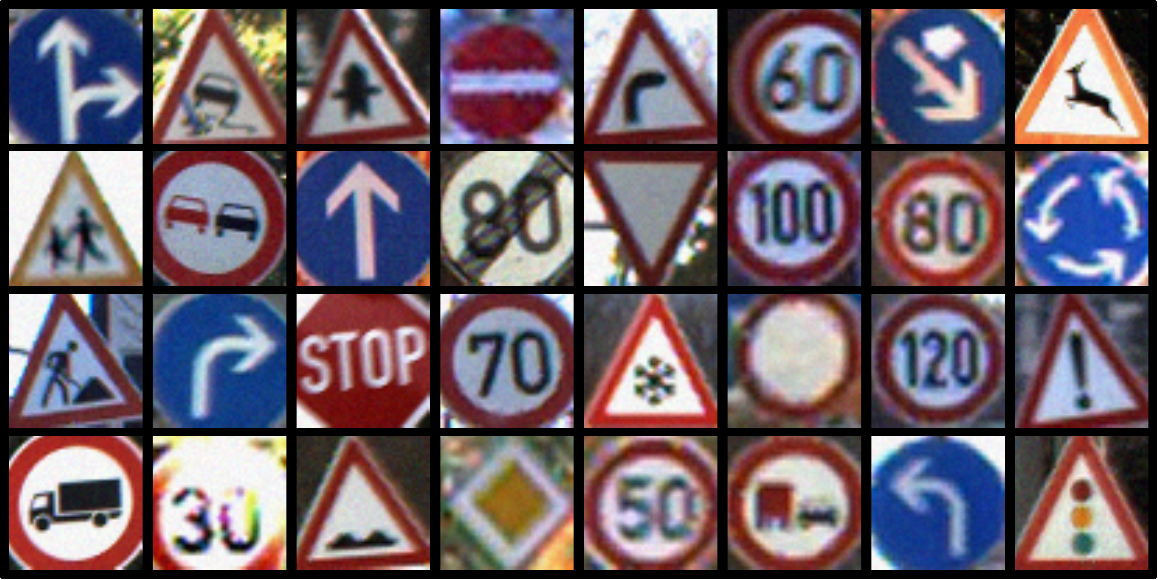}
\caption{The 32 traffic signs learned, one image per class.}
\label{fig:examples_traffic_signs}
\end{center}
\end{small}
\end{figure}

The images are represented as $48\times48$-pixel color (RGB) images (see Figure~\ref{fig:examples_traffic_signs}). We use only 32 out of 43 traffic signs with the most samples. 
For the training data, we use the same number of samples for each class (traffic sign), namely 2,160 of them, making a total of 69,120 images. To reach 2,160 samples per class, images of some classes are used up to 6 times (since the database is unbalanced).
The images used for training are distorted by a random rotation $r$ of $-3.15 \le r \le 3.15$ degrees, horizontal and vertical translations $\Delta_{x}$, $\Delta_{y}$ with $-1.73 \le \Delta_{x}, \Delta_{y} \le 1.73$ pixels, and a scaling factor $s$ with $0.91 \le s \le 1.09$.
The purpose of the distortion is to improve generalization and provide invariances to small misalignments.
We use the official test data, which ensures that the images originate from physical signs different from the ones used for training. The test data consists of 9,030 undistorted images.

We used a simple (non-hierarchical) GSFA architecture, in which PCA is applied first to reduce the dimensionality to $120$ principal components. Afterwards, quadratic GSFA is applied using different training graphs, described below. 
Finally, since this is a classification problem, a nearest centroid classifier is used instead of linear scaling.

The ELL  method is used to construct two training graphs with binary target labels (i.e, a label is either $1$ or $-1$). The first one has 5 labels (compact+5) and the second one has 31 (compact+31). The target labels are defined in Table~\ref{tab:compact_classes}. 
Notice that the first 5 labels (for both graphs) suffice, in principle, to fully code the class information. 

For the compact+5 graph, identical eigenvalues ($\lambda^1_1=\lambda^1_2=\lambda^1_3=\lambda^1_4=\lambda^1_5=0.2$) are used to express equal importance of the target labels.
The compact+31 graph has been included to show the effect of auxiliary labels $\Bell_6, \Bell_7, \dots, \Bell_{31}$.
For this graph, the first five eigenvalues $(\lambda^2_1, \lambda^2_2, \dots, \lambda^2_{5}) = (0.056, 0.056, \dots, 0.056)$ are identical, but the rest decrease linearly: $(\lambda^2_6, \lambda^2_7, \dots, \lambda^2_{31}) = (0.053, 0.051, \dots, 0.002)$, where only three decimal places are shown. 
Thus, the importance given to the auxiliary labels decreases from $\Bell_6$ to $\Bell_{31}$. 
For both graphs, we scale the eigenvalues to make their sum equal to 1.

\begin{table}
\small
\begin{center}
\begin{tabular}{c|c|c|c|c|c|c|c|c|c|c|c|c|c|c|c|c}
$c \rightarrow$ & 1  &  2 &  3 &  4 &  5 &  6 &  7 &  8 &  9 & \dots & 16 & 17 & \dots & 30 & 31 & 32\\
\hline
$\Bell_1(c)$    & -1 & -1 & -1 & -1 & -1 & -1 & -1 & -1 & -1 & \dots & -1 &  1 &  \dots &  1 &  1 &  1\\ 
$\Bell_2(c)$    & -1 & -1 & -1 & -1 & -1 & -1 & -1 & -1 &  1 & \dots &  1 & -1 & \dots &  1 &  1 &  1\\ 
$\Bell_3(c)$    & -1 & -1 & -1 & -1 &  1 &  1 &  1 &  1 & -1 & \dots &  1 & -1 & \dots &  1 &  1 &  1\\ 
$\Bell_4(c)$    & -1 & -1 &  1 &  1 & -1 & -1 &  1 &  1 & -1 & \dots &  1 & -1 & \dots & -1 &  1 &  1\\ 
$\Bell_5(c)$    & -1 &  1 & -1 &  1 & -1 &  1 & -1 &  1 & -1 & \dots &  1 & -1 & \dots & 1 &  -1 &  1 \\ 
\hline
$\Bell_6(c)$    & -1 &  1 &  1 & -1 &  1 & -1 & -1 &  1 &  1 & \dots & -1 &  1 & \dots & -1 & -1 &  1 \\ 
$\Bell_7(c)$    & -1 & -1 &  1 &  1 &  1 &  1 & -1 & -1 &  1 & \dots &  1 &  1 & \dots &   1 & -1 & -1 \\ 
\vdots&\vdots&\vdots&\vdots&\vdots&\vdots&\vdots&\vdots&\vdots&\vdots&\vdots&\vdots&\vdots&\vdots&\vdots&\vdots&\vdots \\
$\Bell_{30}(c)$ & -1 &  1 & -1 &  1 &  1 & -1 &  1 & -1 & -1 & \dots & -1 & -1 & \dots &  -1 & 1 & -1 \\ 
$\Bell_{31}(c)$ & -1 &  1 &  1 & -1 & -1 &  1 &  1 & -1 & -1 & \dots & -1 & -1 & \dots &  1 & 1 & -1 \\ 
\end{tabular}

\caption{\label{tab:compact_classes}
Target labels used to code the class information, compactly expressed as a function of the class number $c$. 
The compact+5 graph is constructed with labels $\Bell_1$ to $\Bell_5$, whereas the compact+31 graph with $\Bell_1$ to $\Bell_{31}$. The first five labels can be seen as the original ones and the rest as auxiliary.
}
\end{center}
\end{table}

We choose $C=2^5$ classes, because powers of two make it simple to obtain binary labels with a weighted zero mean, weighted unit variance, and weighted decorrelation, as follows. 
The first five original labels can be computed as
$\Bell_j(c)=2 (\frac{c-1}{2^{5-j}} \mod 2)-1$, where the division is integer division and ``$\ttxt{mod}$'' is the modulo operation.
The auxiliary labels are computed as the product of two or more labels $\Bell_1$ to $\Bell_5$, possibly multiplied by a factor $-1$ to make the label assigned to the first class  negative.
More concretely, $\Bell_6$ is the product of all original labels, $\Bell_7$ to $\Bell_{11}$ are all the products of four of them, $\Bell_{12}$ to $\Bell_{21}$ are all the products of three, and $\Bell_{22}$ to $\Bell_{31}$ are all the products of two (e.g., $\Bell_6 = \Bell_1 \Bell_2 \Bell_3 \Bell_4 \Bell_5$, $\Bell_7 = -\Bell_1 \Bell_2 \Bell_3 \Bell_4$, $\Bell_8 = -\Bell_1 \Bell_2 \Bell_3 \Bell_5$, $\Bell_{30} = -\Bell_3 \Bell_5$, $\Bell_{31} = -\Bell_4 \Bell_5$).
 
For both graphs, we set $\vect{v}=\vect{1}$. The corresponding eigenvectors are $\vect{u}_j \stackrel{(\ref{eq:u_in_ell})}{=} Q^{-1/2} \Bell_j$, where $Q \stackrel{(\ref{eq:R_Q_long})}{=} N \cdot 1 = 69$,$120$ ($N$ is the number of training images). These eigenvectors are also binary and allow for a fast computation of the covariance matrix in $\bigO(LNI^2+I^3)$ operations, where $L$ is the number of target labels.

\begin{figure}
\begin{small}
\begin{center}
\includegraphics[width=0.6\textwidth]{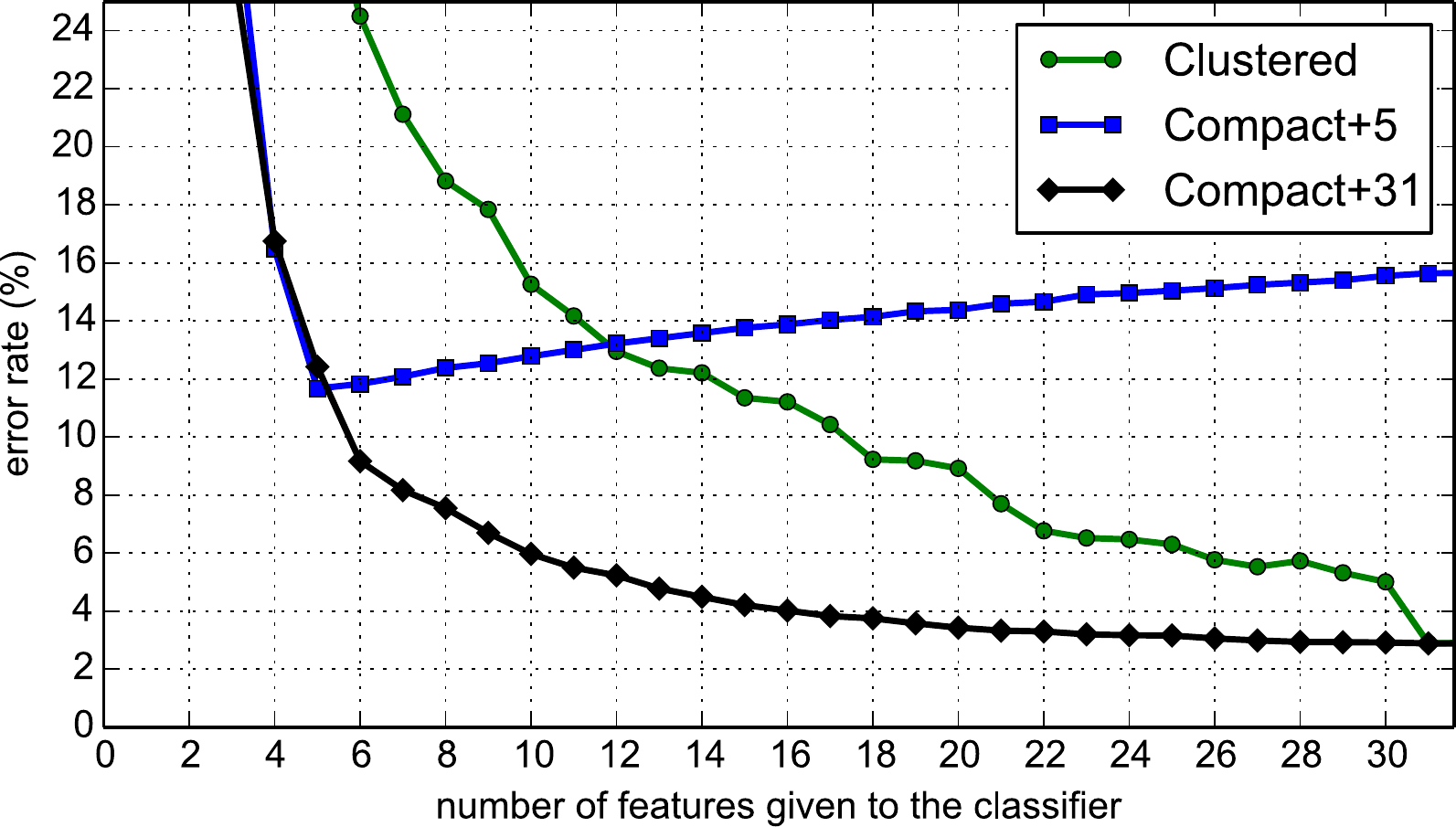}
\small
\caption{Classification error when GSFA is trained with the compact+5, compact+31 and clustered graph (FDA). This error is a function of the graph and the number of slow features $d$ passed to the classifier. 
For the clustered graph, dropping even a single feature might increase the error rate significantly. For instance, the error rate of using 30 features computed with the clustered graph is worse than the error rate of 13 features computed with the compact+31 graph. 
Performance on 9,030 test samples, averaged over 10 runs (largest standard deviation 0.15\%).
}
\label{fig:compact_classes}
\end{center}
\end{small}
\end{figure}

The classification error is plotted in Figure~\ref{fig:compact_classes}, where the number of slow features $d$ given to a nearest centroid classifier ranges from 4 to 31. For comparison, the clustered graph is also evaluated.
For $d=5$ features, the compact+5 graph results in the best accuracy with an error rate of 11.67\%, against 12.42\% (compact+31) and 29.74\% (clustered).
However, the error rate of the compact+5 graph increases if one preserves more than 5 features, indicating that additional features contain little or no discriminative information.
For $6 \le d \le 30$, the compact+31 graph yields clearly better accuracy than the other graphs.
Interestingly, for $d=31=C-1$ features, the compact+31 and clustered graph give identical error rates of $2.89\%$, which is their top performance. In this case, the features extracted are \emph{different} but contain the \emph{same information} since they can be mapped to each other linearly.
In other words, the first $31$ free responses of both graphs describe the same subspace. Any \emph{single} slow feature from the compact+31 graph contains ideally 1 bit of discriminative information (which might be redundant to the others). In contrast, the first features extracted by the clustered graph might sacrifice discriminative information to minimize within-class variance (e.g., a feature $\vect{y}(c)=\big( (\frac{C}{2})^{1/2}, -(\frac{C}{2})^{1/2}, 0, 0, \dots, 0 \big)$ has minimal (zero) within-class variance but provides little discriminative information, because only the first two classes can be identified from it). 
Using $d>31$ features does not improve accuracy in any case.

One assumption of this method is that the feature space is complex enough to allow the extraction of features that approximate the binary labels. If the feature space is poor, the compact graphs might not bring any advantage over the clustered one.

The results suggest that if the number of features to be preserved $d<C-1$ is known, one might improve accuracy by creating a graph that uses exactly $d$ target labels.
If the number of output features is unclear, a graph coding several auxiliary labels with decreasing eigenvalues, e.g.\ compact+$(C-1)$, might provide better robustness (see Figure~\ref{fig:compact_classes}). 

\section{Discussion}
\label{sec:discussion_ELL}
In this article, we propose exact label learning (ELL) for the construction of training graphs, in which the final label estimation is just a linear transformation of the slowest feature extracted.
The method allows the direct solution of regression problems with GSFA,  without having to recur to a supervised post-processing step. In other words, given a new input sample (e.g.\ an input image) the first feature computed using an ELL graph directly provides an approximation of the label (or a linear transformation of it).
In practice, even better results may be achieved using more than one feature and supervised post-processing.

It is crucial to emphasize that GSFA optimizes feature slowness, which depends on the particular training graph used, and not label estimation accuracy.
However, when the ELL method is used, the training graphs define a slowness objective that requires optimizing an output similarity function, where the similarities are intimately related to the desired label similarities.
As a result, the feature slowness and estimation accuracy objectives become equivalent when 
$\mathcal{F}$ becomes unlimited. That is, the slowest possible features that can be extracted (i.e.\ optimal free responses) are equal to a normalized version of the label(s).
In practice, $\mathcal{F}$ is finite to allow generalization from training to test data and, if the features extracted are slow enough (i.e.\ close to the optimal free responses), they are also good solutions to the original regression problem.

Supervised learning problems on high-dimensional data are of great practical importance, but they frequently result in systems with large computational demands. A common approach is to apply feature extraction, dimensionality reduction, and a supervised learning algorithm.
A promising alternative approach is hierarchical GSFA (HGSFA), because its complexity scales in some cases even linearly w.r.t.\ the input dimensionality and the number of samples.
Furthermore, when HGSFA is trained with an ELL graph, the resulting architecture is simple and homogeneous, as shown in Figure~\ref{fig:ELLidea}.c.

We have proved the usefulness of the ELL method by showing three types of applications that are relevant in practice: ELL regression with multiple labels, analysis of training graphs, and classification with compact discriminative features.

The results show that encoding auxiliary labels derived from the original one (e.g.\ ``higher-frequency'' transformations of it) improve performance. This is particularly relevant for cascaded or convergent hierarchical GSFA networks.
The auxiliary labels contain information that (partially) determines the original label. 
A GSFA node that receives this information might be able to extract the original label more accurately than without it.

Multiple labels can be learned simultaneously, for instance to code different aspects of the input at once (e.g.\ object color, size, shape, orientation).
The use of multiple labels is inspired by biological systems, where complementary information channels have been observed and might improve feature robustness, for example under incomplete information~\citep{KruegerJanssenEtAl-2013}.
Learning gender and color simultaneously yielded clearly smaller estimation errors than when gender and color were estimated separately. This shows that multiple label learning is not only theoretically possible, but that coding complementary information channels might boost accuracy in practice. 
For instance, an automatic system for face processing might benefit from the simultaneous extraction of the subject's identity, age, gender, race, pose, and expression.



The experiments on gender (and skin-color) estimation from artificial face images verify that the ELL method works in practice.
The complexity of training a single GSFA node with an ELL graph is $\bigO(IN^2+I^2N+I^3)$ operations, where $I$ is the input dimensionality (possibly after a nonlinear expansion), and $N > I$ is the number of samples.
%
%
For comparison, the serial graph has a complexity of $\bigO(I^2N+I^3)$. Thus, the main limitation of using ELL graphs is the training complexity.

The analytical and practical results show the strength of the serial graph when only a single label is available. In this case, the ELL graph provided marginally better estimations than the serial one (an RMSE of 0.345 with the $\ttxt{ELL}^g$-$40$ graph vs. 0.349 with the serial graph, in both cases using 3 features and the soft GC post-processing method), but the computation time was 25 times larger.  
Although the shape of the slowest feature extracted with the serial graph might be less similar to the label, a monotonic transformation of the slowest feature learned by a nonlinear supervised step (e.g.\ soft GC) might suffice to approximate it.

However, the results show that if two or more (intrinsically connected) labels are available, the accuracy of using ELL graphs may further increase. Efficient pre-defined graphs are not available in this case. In the gender estimation experiment, the RMSE was improved to 0.277 by jointly learning gender and skin ($\ttxt{ELL}^{g,c}$-$(2 \times 5)$ graph, 3 features, soft GC). Hence, a particularly promising application for the ELL method is multiple label learning.

Various methods for mapping the slowest feature to a label were tested.
The linear scaling method is interesting from a theoretical point of view. However, as one would expect, it provided worse accuracy for test data than the soft GC method, which is nonlinear and supervised. Therefore, the latter might be preferred in practical applications.
Moreover, in this scenario, supervised post-processing methods might be computationally inexpensive, because their input is frequently low-dimensional (e.g., we used 1 to 3 slow features for gender estimation).




Although ELL was originally designed for regression, we show that it can also be useful for classification when particular labels are learned.
The experiment on traffic sign classification shows the benefit of using compact discriminative features, implemented here by learning multiple binary labels. 
The resulting system has a much smaller classification error than the clustered graph (equivalent to nonlinear FDA) when the number of output dimensions is less than $C-1$, where $C$ is the number of classes.
One can combine the method with HGSFA for classification with many classes, coding the discriminative information in much fewer than $C-1$ features and reducing the number of signals computed in the network, which might also reduce overfitting. 
Although ideally $\log_2(C)$ binary target labels suffice for perfect classification, the experiments show that additional target labels via auxiliary labels improve classification accuracy in practice.

Interestingly, the clustered graph for $C$ classes (equivalent to FDA) and the compact+$(C-1)$ graph are equivalent if the latter is constructed with constant positive eigenvalues $\lambda_1 = \cdots = \lambda_{C-1} = 1/(C-1)$. 
The reason for this equivalence is that this compact+$(C-1)$ graph would only have 
within-class transitions, because transitions between different classes cancel out each other. Therefore, the clustered graph can be seen as a special case of the compact+$(C-1)$ graph, with maximum label redundancy ($C-1$ target labels) and giving equal importance (eigenvalues) to all of them.



Future work in the scope of ELL might include the construction of training graphs that are efficient (like the serial one) but at the same time offer the versatility of the ELL graphs.
For example, up to now, efficient (pre-defined) training graphs can only learn a single target label (and its higher-frequency harmonics). 
Therefore, we are interested in designing {\it efficient} graphs that allow learning multiple labels.

Another possible application of ELL is the generation of (partially) sparse feature representations. For instance, for $C=8$ classes, one might learn features 
$\Bell_1(c) = (2,-2,0,0,0,0,\\0,0)$, $\Bell_2(c) = (0,0, 2,-2,0,0,0,0)$, $\Bell_3(c) = (0,0,0,0,0,0,2,-2,0,0)$, $\Bell_4(c) = (0,0,0,0,0,0,\\2,-2)$, $\Bell_5(c) = (2^{1/2},2^{1/2},-2^{1/2},-2^{1/2},0,0,0,0)$, $\Bell_6(c) = (0,0,0,0,2^{1/2},2^{1/2},-2^{1/2},-2^{1/2})$, and $\Bell_7(c) = (1,1,1,1,-1,-1,-1,-1)$. 
These features have zero mean, unit variance, and are decorrelated. Furthermore, $\Bell_1$ to $\Bell_4$ make the smallest possible contribution to the $L_1$ norm, thus this construction is greedy (it might not minimize the $L_1$ norm globally).

An open research question is how to choose the auxiliary labels. It is clear they should depend on the original label(s), but it is unclear how to compute them to maximize estimation accuracy, how many of them should be coded, and which eigenvalues should be used.
For classification with $C=32$ classes, \emph{linearly} decreasing eigenvalues (for the auxiliary labels) provided great results, but other eigenvalues might be better if $C$ is very large.



Hierarchical processing and the slowness principle are two powerful brain-inspired learning principles. 
The strength of SFA originates from its theoretical foundations in the field of learning of invariances and the generality of the slowness principle.
For practical supervised learning applications, HGSFA provides good accuracy and efficiency and still profits from strong theoretical foundations.
An advantage of relying on such general principles is that the resulting algorithms are application independent and not confined to a particular problem or input feature representation. 
Of course, fine tuning the network parameters and the integration of problem-specific knowledge are always possible for additional performance.
The proposed method explores the limits of HGSFA and is valuable as a theoretical tool for the analysis and design of training graphs.
However, the results show that with certain adaptations (e.g.\ the use of supervised post-processing) it is also sufficiently robust to be applied to practical computer vision and machine learning tasks.


\vskip 0.2in
\small
\bibliography{references}
\end{document}